\UseRawInputEncoding

\documentclass[10pt,twocolumn,letterpaper]{article}

\usepackage[pagenumbers]{cvpr} 

\definecolor{cvprblue}{rgb}{0.21,0.49,0.74}
\usepackage[pagebackref,breaklinks,colorlinks,allcolors=cvprblue]{hyperref}

\usepackage{xcolor}
\usepackage{soul}

\usepackage{tocloft}
\sethlcolor{yellow!80} 

\def\paperID{9973} 
\def\confName{CVPR}
\def\confYear{2026}

\title{Beyond Binary Preference: Aligning Diffusion Models to Fine-grained Criteria by Decoupling Attributes}

\author{Chenye Meng$^1$
\and Zejian Li$^1$ \and Zhongni Liu$^2$ \and Yize Li$^1$\and Changle Xie$^1$ \and Kaixin Jia$^1$ \and Ling Yang$^3$ \and Huanghuang Deng$^1$ \and Shiying Ding$^1$ \and Shengyuan Zhang$^1$ \and  Jiayi Li$^4$ \and  Lingyun Sun$^1$ \\
{\small $^1$ Zhejiang University \quad \small $^2$ University of Electronic Science and Technology of China‌}\\{ \small $^3$ Peking University \quad \small $^4$ University of Nottingham Ningbo China} \\
{\tt\small $^1$ \{zejianlee,mengcy\}@zju.edu.cn}\\
}

\begin{document}
\maketitle

\begin{abstract}
Post-training alignment of diffusion models relies on simplified signals, such as scalar rewards or binary preferences.
This limits alignment with complex human expertise, which is hierarchical and fine-grained. 
To address this, we first construct a hierarchical, fine-grained evaluation criteria with domain experts, which decomposes image quality into multiple positive and negative attributes organized in a tree structure.
Building on this, we propose a two-stage alignment framework. First, we inject domain knowledge to an auxiliary diffusion model via Supervised Fine-Tuning. Second, we introduce Complex Preference Optimization (CPO) that extends DPO to align the target diffusion to our non-binary, hierarchical criteria. Specifically, we reformulate the alignment problem to simultaneously maximize the probability of positive attributes while minimizing the probability of negative attributes with the auxiliary diffusion.
We instantiate our approach in the domain of painting generation and conduct CPO training with an annotated dataset of painting with fine-grained attributes based on our criteria.
Extensive experiments demonstrate that CPO significantly enhances generation quality and alignment with expertise, opening new avenues for fine-grained criteria alignment.

\end{abstract} 

\begin{figure}[t]
    \centering
    \includegraphics[width=1\linewidth]{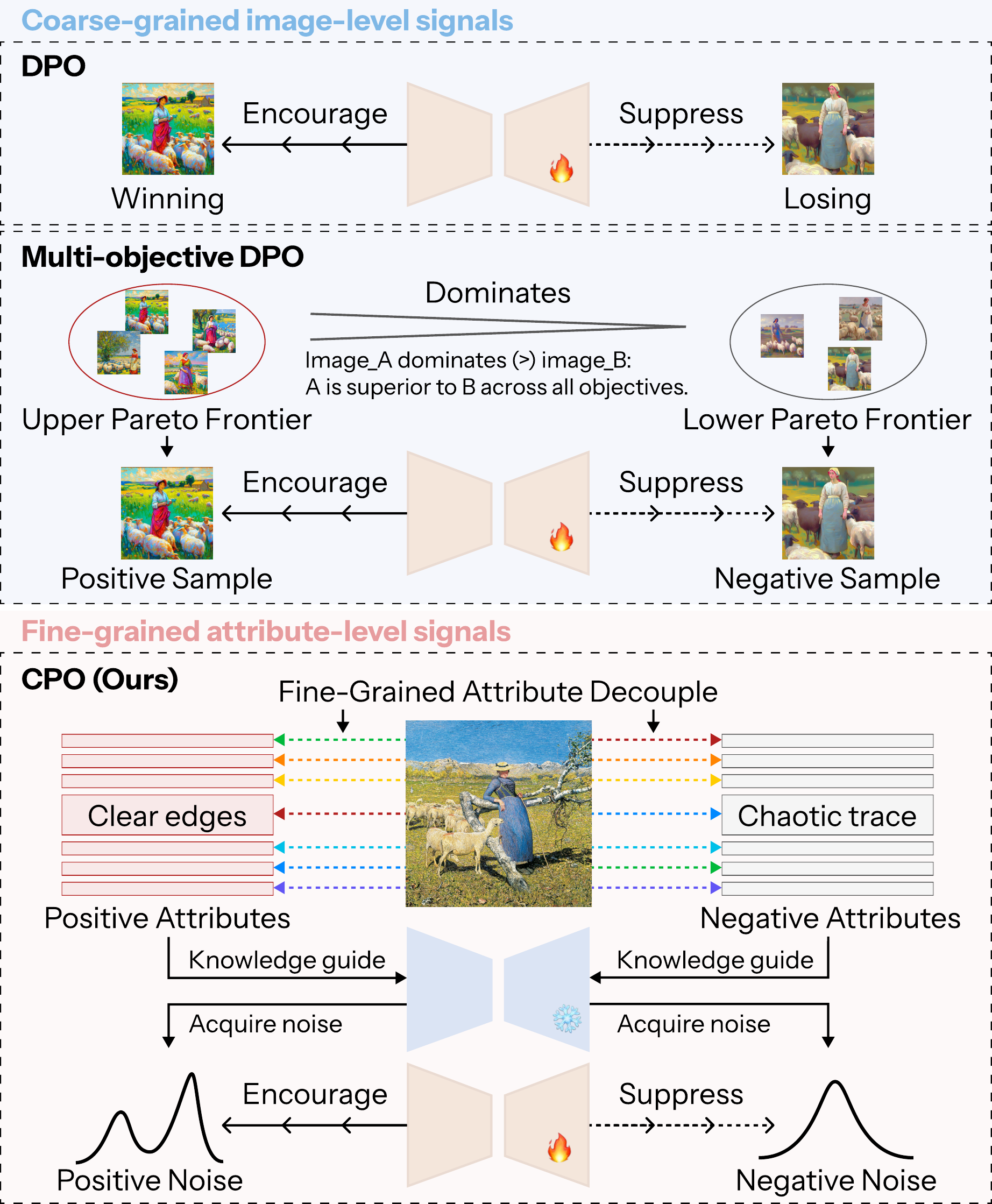} 
    \caption{Existing methods rely on coarse-grained, scalar or binary image-level reward signals. In contrast, our method leverages human expert knowledge for fine-grained attribute decoupling, guiding the model directly from the noise space to approach positive and avoid negative directions.}
    \label{fig:teaser}
\end{figure}


\section{Introduction}
In the new era of generative AI, ``evaluation has become more important than training''~\cite{Yao_AISecondHalf_2025}. 
The quality and nature of evaluation and data fundamentally define the upper limit of a model's capabilities. 
Recent post-training strategies, such as Reinforcement Learning from Human Feedback (RLHF) including DPO~\cite{rafailov2023directDPO} and GRPO~\cite{shao2024deepseekmath}, have demonstrated significant efficacy in enhancing generative models. However, these prevailing frameworks fundamentally depend on the scores of reward models or binary human preferences of winning and losing samples they are optimized for (\cref{fig:teaser}). Such simplified, coarse evaluation criteria lead to a substantial gap when compared to the complex and nuanced patterns of human cognition in real world.

Human evaluation does not follow such a uni-dimensional or regularized process. Consistent with existing research, we summarize three features of human expert evaluation: (1) Multi-dimensional, assessing multiple dimensions simultaneously (such as composition, color relations and brushwork in paintings); (2) Discrete, employing symbolic labels rather than continuous scores; and (3) Non-equilibrium, meaning the applicable set of evaluation labels dynamically shifts with samples. 

This highlights a critical insight: positive ($A_{pos}$) and negative ($A_{neg}$) attributes are not merely opposites. 
Their relationship is complex. They may be mutually exclusive in some cases, while in others they can coexist within the same sample. Existing post-training frameworks, which typically optimize a single utility function, are ill-equipped to process such complex signals. We argue that an evaluation paradigm aligned with fine-grained human cognition can provide more specific, interpretable guidance, leading to enhanced generation quality and controllability.

To bridge this chasm, we go beyond binary preferences and propose a new evaluation paradigm. We construct a hierarchical, multi-dimensional evaluation 
criterion with domain experts. We instantiate our approach in the domain of painting generation, developing a domain-specific knowledge system comprising 7 root dimensions (e.g., Composition, Color Relations) and 246 pairs of positive/negative attributes. To operationalize this system, we build a domain-expert agent that annotates 10,277 collected images of paintings, transforming expert evaluation into discrete, symbolic semantic labels that explicitly identify coexisting positive and negative attributes ($A_{pos}, A_{neg}$).

Building on this fine-grained feedback, we propose a novel two-stage post-training strategy. In the first stage, we inject domain knowledge into a pre-trained model via Supervised Fine-Tuning, yielding an expert model $\theta_1$ sensitive to these complex attributes. In the second stage, we introduce Complex Preference Optimization (CPO), a novel preference learning algorithm to train the final generative model with decoupling attributes learned in the expert model and. 
Given a noisy sample from the training set, $\theta_1$ provides an ideal noise prediction $z^w$ (winner) mainly conditioned on $A_{pos}$ and non-ideal $z^l$ (loser) mainly on $A_{neg}$. By assuming the winner prediction guides the noisy training sample to a winning output and vice versa, we perform a preference optimization that steers the final trained model toward $A_{pos}$ yet away from $A_{neg}$. 
In this case, the trained model generates images aligned with domain-specific evaluation criteria given only the content prompt without specified complex positive attributes. 


In practice, we observe instability of preference optimization and propose a new stabilizing strategy.
The instability is manifested by that the term on losing samples dominates the training while that on winning samples fails to converge consistently.
We attribute this phenomenon to the behavior of minimizing a negative squared error, and thus propose a new stretegy that translates the loss term for the losing samples. The translation restricts the norm of backward gradients but remain the gradient direction as the original loss.
Our strategy encourages a balance between the gradients of winning and losing samples.

Extensive experiments demonstrate that our approach significantly enhances generation quality and alignment with expert preferences. 
Our stabilizing strategy boosts training by over 10 times faster compared to the counterpart with the original loss.
Our work validates the merit of fine-grained evaluation and sheds light on future post-training paradigms.
In summary, our contributions are as follows:
\begin{itemize}
    \item We extend the simplified binary preferences and propose a new, human-aligned evaluation criteria based on multi-dimensional, discrete, and non-equilibrium expert criteria. We instantiate this criterion and develop a ``domain-expert agent" to create a fine-grained dataset with positive and negative attributes.
    \item We propose a novel two-stage post-training strategy, dubbed Complex Preference Optimization (CPO), which aligns a diffusion model by decoupling the positive and negative attributes inside generated samples.
    \item We introduce a new stability strategy, resolving optimization instabilities by balancing gradients from the postive and negative samples.
\end{itemize}

\section{Related Work}
\textbf{Preference optimization dataset.} The efficacy of preference alignment is constrained by the feedback signal's granularity. Foundational datasets, including Pick-a-Pic~\cite{Yuval2023Pick-a-Pic}, ImageReward~\cite{xu2024ImageReward},HPS~\cite{wu2023hpsv2, ma2025hpsv3}, and LAION-Aesthetic~\cite{Schuhmann2022LAION_5B} establish the field by collecting large-scale binary preferences (winning/losing) or monolithic aesthetic scores (e.g., 1-10). However, these simplified evaluation criteria result in a pronounced discrepancy between the feedback signal and the complex, fine-grained human evaluation. This limitation is gaining recognition, evidenced by the emergence of RichHF-18k~\cite{liang2024richHF} and VisionReward~\cite{xu2024visionreward}. They assess human preferences along multiple dimensions, yet the evaluation remains at a coarse level.

\textbf{Direct preference optimization.}
Traditional Reinforcement Learning from Human Feedback (RLHF)~\cite{bai2022traininghelpfulharmlessassistant,Ouyang2022TraininglanguageHF} typically requires the explicit training of a reward model~\cite{Uehara2024Feedback,Fan2023DPOK,black2024DDPO,Eyring2024Reno,zhang2025itercomp}. To reduce the overhead, Direct Preference Optimization (DPO)~\cite{rafailov2023directDPO} is introduced for language models as a stable, RL-free objective, which is successfully adapted to vision by Diffusion-DPO~\cite{wallace2024diffusionDPO}. Subsequent studies primarily focus on refining the optimization process rather than the feedback signal itself. This includes process-guided and step-supervised methods such as SPO~\cite{liang2025aestheticSPO}, D3PO~\cite{yang2024D3PO}, and A Dense Reward View~\cite{yang2024adensereward}; inversion-based approaches such as Inversion-DPO~\cite{li2025inversionDPO} and InPO~\cite{lu2025inpo} that enable efficient latent tuning; and trajectory-level optimization methods such as Diffusion-Sharpening~\cite{tian2025diffusionsharpening}. Recently, Negative Preference Optimization (NPO)~\cite{zhang2024NPOlanguage} is explored for unlearning bad concepts in language models. Building upon this idea, Diffusion-NPO~\cite{wang2025diffusionnpo} and Self-NPO~\cite{wang2025selfNPO} extend the framework to the visual domain by explicitly training a negative preference model on switched data pairs. Nevertheless, these methods are all based on coarse-grained scalar or binary reward, and some require the training of an auxiliary negative preference model.


\textbf{Multi-objective optimization.}
Recent research addresses the ``one-preference-for-all” problem by advancing toward multi-objective optimization, which aims to balance conflicting monolithic rewards. In language modeling, MODPO~\cite{zhou-etal-2024-beyondMODPO} produces a Pareto front of models trading off objectives such as helpfulness and harmlessness. This paradigm is extended to vision by CaPO~\cite{lee2025calibrated_capo}, which aligns diffusion models with multiple distinct rewards. Parrot~\cite{lee2024parrot} and Preference-Guided Diffusion~\cite{annadani2025preferenceguided} also pursue Pareto-optimal solutions. However, they operate at an aggregated reward to balance different rewards and thus fail to exploit fine-grained attribute information within images.


\section{Domain-specific Fine-grained Evaluation}
\label{sec:Domain-specific Fine-grained Evaluation}
Prevailing preference optimization frameworks~\cite{christiano2017deep, lee2023aligning, rafailov2023directDPO, wallace2024diffusionDPO} are founded on simplified evaluation paradigms. They collapse complex, multi-dimensional human evaluation into a uni-dimensional signal, such as a scalar reward or a binary preference. This simplification widens the chasm between the simplified feedback and the granular, complex nature of real-world human cognition~\cite{freedman2001categorical, treisman1980feature, leder2004model}. This fundamental limitation of the signal structure inherently restricts the potential for fine-grained model improvement. 

To bridge this chasm, we first develop a new evaluation paradigm imitating expert evaluation.
We choose painting generation as our focused domain but our proposed paradigm and method can be easily extended to other scenarios without loss of generality.
Collaborating with painting experts, we construct a 5-level knowledge hierarchy for evaluation, which comprises 7 root dimensions (including Composition, Color Relations, etc.) and 246 manually-defined, well-organized pairs of positive/negative attributes. Please refer to our SM for details.

We reveal that human evaluation has three features. (1) The evaluation is Multi-dimensional, and experts assess multiple attributes simultaneously.
Notice that each of our 7 root dimensions has separate multi-level sub-dimensions to organize attributes. The Composition defines composition category, visual guidance, image richness, visual equilibrium and visual rhythm as sub-dimensions. Again, each sub-dimension has its own children dimensions. Therefore, Multi-Dimension here is also hierarchical.
(2) The evaluation language is Discrete; experts tend to employ multiple attributes rather than continuous scores for fine-grained evaluation.
(3) The evaluation is Non-Equilibrium, and the applicable set of attributes dynamically shifts with the image's content and style. 
For example, in the sub-dimension of composition category, we have composition of symmetry, asymmetry and geometry as children dimensions. 
One painting may be of axis-symmetric as a leaf attribute of symmetry and also of circular composition as in geometry. However, the painting may fail to break the shape of the circle and thus suffer from a negative attribute of `close circle without shape breaking'.
Another painting may be center-symmetric and of radial composition simultaneously, while it may suffer from another negative attribute of `ambiguous center' because the center to display radial composition is not clear enough.
This example shows the applicable attributes vary across different samples (non-equilibrium).

Two phenomena pose new challenges to existing post-training methods. First, negative ($A_{neg}$) attributes co-exist with the positive ($A_{pos}$) in one single painting sample. This requires a post-training method to decouple attributes in samples.
Second, positive attributes ($A_{pos}$) can be mutually exclusive when they share the same penultimate sub-dimension. An example is a painting cannot be of upward triangle and circular symmetry simultaneously, and both kinds of symmetry share the same ancestor as geometric composition. 
This means learned positive attributes vary in each training sample.
Existing post-training frameworks to optimize a single or multiple utility functions are ill-equipped to process such complex, multi-faceted signals. 

To operationalize this nuanced understanding, we introduce a domain-expert agent that employs a ``Deconstruct-Structure-Quantify'' paradigm. This agent leverages our hierarchical knowledge framework—structured as a 5-level tree with 7 root dimensions---to mimic expert evaluation into prompts. The terminal nodes represent discrete, symbolic semantic labels, explicitly identifying both positive and negative attributes. 
This structure facilitates non-equilibrium evaluation: rather than applying a universal metric, the agent dynamically activates a relevant subset of attributes from this extensive knowledge base tailored to the specific image. 
Utilizing this agent, we annotated 10,277 paintings, creating a domain-specific dataset $D = \{(x_0, y, A_{pos}, A_{neg})\}$, where $x_0$ is an image of one painting, $y$ its prompt, and $A_{pos}$ and $A_{neg}$ are the sets of positive and negative attributes assigned by the domain-expert agent. By manual investigation, the annotation accuracy is acceptable. Please see SM for more details.

\begin{figure*}[t]
    \centering
    \includegraphics[width=1\linewidth]{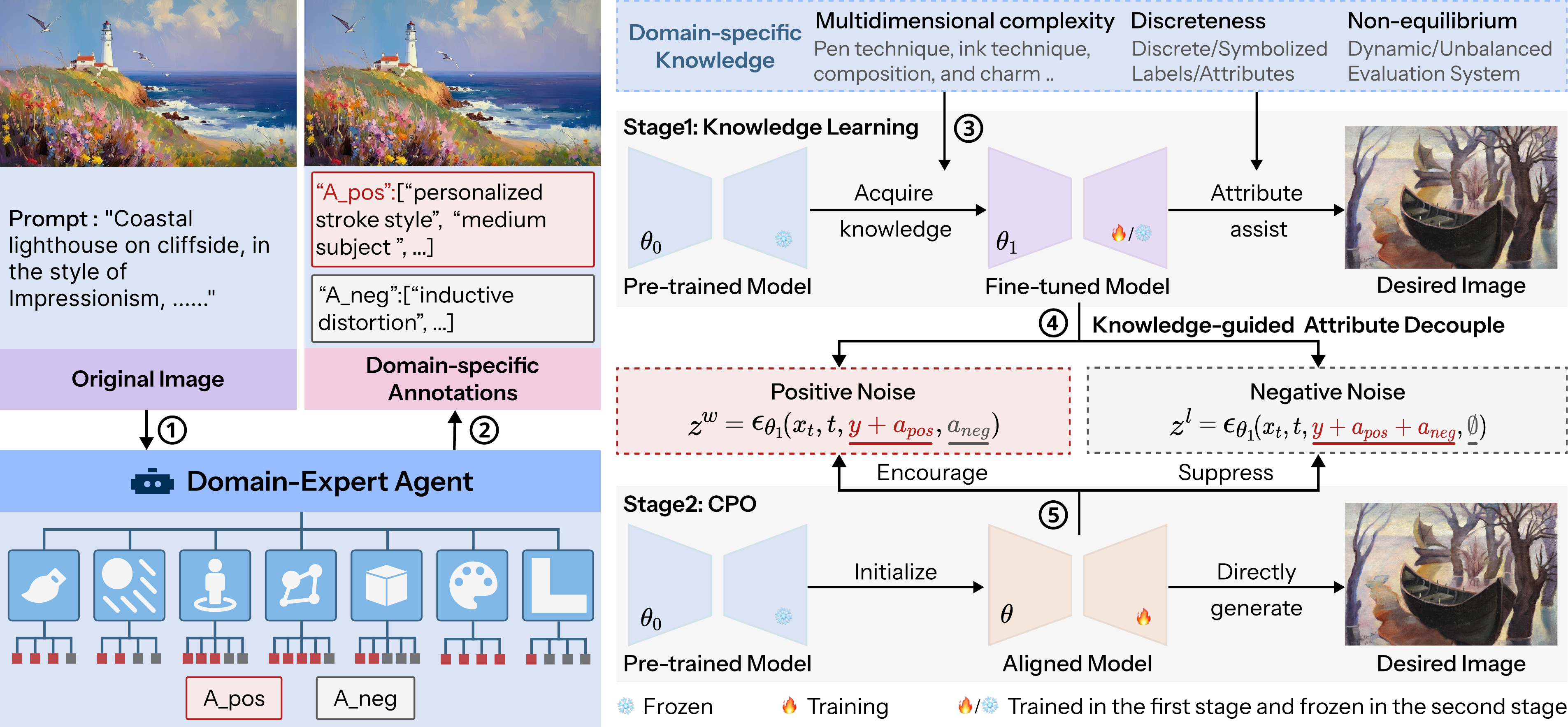} 
    
    \caption{
        The pipeline of our framework. 
        The Domain-Expert Agent decomposes image along 7 dimensions, which are represented as: 
        %
        %
        \protect\raisebox{-0.5ex}{\protect\includegraphics[height=2.5ex]{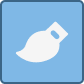}} \textbf{Brushstroke and Texture},
        \protect\raisebox{-0.5ex}{\protect\includegraphics[height=2.5ex]{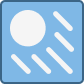}} \textbf{Light and Shadow},
        \protect\raisebox{-0.5ex}{\protect\includegraphics[height=2.5ex]{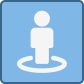}} \textbf{Shape and Posture},
        \protect\raisebox{-0.5ex}{\protect\includegraphics[height=2.5ex]{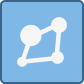}} \textbf{Composition},
        \protect\raisebox{-0.5ex}{\protect\includegraphics[height=2.5ex]{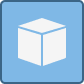}} \textbf{Perspective and Space},
        \protect\raisebox{-0.5ex}{\protect\includegraphics[height=2.5ex]{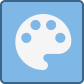}} \textbf{Color relationship},
        and \protect\raisebox{-0.5ex}{\protect\includegraphics[height=2.5ex]{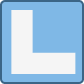}} \textbf{Edge relationship}.
        %
        %
        Notice that the visualization of the attribute hierarchy in the agent is simplified. 
        The full hierarchy is of 5 levels with 246 attribute pairs in the leaf nodes. Post-annotation, we first conduct SFT to obtain the model $\theta_1$. This model is then used to dynamically acquire noise signals that aggregate decoupled attribute information. Subsequently, the aligned model is trained to learn the positive direction while suppressing the negative direction.
    }
    \label{fig:pipeline} 
\end{figure*}

\section{Preliminary}
\textbf{Diffusion models}~\cite{ho2020ddpm, song2020ddim} learn data distributions by reversing a gradual noising process. Given a clean sample $x_0 \sim q(x_0)$, a forward process progressively adds Gaussian noise to produce a sequence $x_{1:T}$ according to
\begin{equation}
    q(x_t|x_{t-1}) = \mathcal{N}(x_t; \sqrt{1-\beta_t}x_{t-1}, \beta_t \mathbf{I}),
\end{equation}
where ${\beta_t}$ controls the noise schedule.
A neural network $\epsilon_\theta(x_t, t, c)$ is trained to approximate the reverse process
\begin{equation}
    p_\theta(x_{t-1}|x_t, c) = \mathcal{N}(x_{t-1}; \mu_\theta(x_t, t, c), \sigma_t^2 \mathbf{I}),
\end{equation}
by predicting the injected noise $\epsilon$ at timestep $t$ with condition $c$. Training minimizes the expected reconstruction error between true and predicted noise, often expressed as
\begin{equation}
\label{eq: loss-diffusion}
    L_{\text{DM}} = \mathbb{E}_{x_0, t, c, \epsilon}\left[\|\epsilon - \epsilon_\theta(x_t, t, c)\|_2^2\right].
\end{equation}
This formulation enables sampling through iterative denoising from pure noise, generating images that are consistent with the given condition.

\textbf{Classifier-Free Guidance (CFG)}~\cite{ho2021classifierfree} is a cornerstone technique in diffusion models for enhancing conditional control during inference without requiring an explicit classifier. The model is trained to learn both a conditional prediction $\epsilon_\theta(x_t, t, c)$ and an unconditional prediction $\epsilon_\theta(x_t, t, \emptyset)$ by randomly dropping the condition $c$ during training.
At inference time, the final noise prediction $\hat{\epsilon}$ is computed by extrapolating from the unconditional baseline in the direction of the conditional semantics:
\begin{equation}
    \hat{\epsilon}(x_t, t, c) = \epsilon_\theta(x_t, t, \emptyset) + \omega \cdot (\epsilon_\theta(x_t, t, c) - \epsilon_\theta(x_t, t, \emptyset))
\end{equation}
where $\omega \geq 0$ is the guidance scale. This structure allows for a trade-off between sample fidelity (to the condition $c$) and diversity. Inspired by this, our work leverages a similar extrapolation structure to guide the diffusion model in generating outputs that align with positive attributes while avoiding negative attributes.

\textbf{Direct Preference Optimization (DPO)}~\cite{rafailov2023directDPO}  reformulates the reward-learning step of RLHF into a direct policy optimization problem. Given preference pairs $(c,x_0^w,x_0^l)$, the Bradley–Terry model~\cite{bradley1952rank} assumes
\begin{equation}
    p(x_0^w \succ x_0^l|c)=\sigma(r(c,x_0^w)-r(c,x_0^l)),
\end{equation}
where $r(\cdot)$ is the latent reward. The standard constrained reward maximization is formulated as
\begin{equation}
    \max_{p_\theta} \mathbb{E}_{x_0 \sim p_\theta}[r(c,x_0)] - \beta \mathbb{D}_{\text{KL}}[p_\theta(x_0|c) \| p_\text{ref}(x_0|c)],
\end{equation}
where the hyperparameter $\beta$ controls regularization. It optimizes a conditional generative distribution $p_\theta$ to maximize the expected reward while regularizing the KL-divergence with respect to a reference distribution $p_\text{ref}$.

Noting that the global optimal policy takes the form $p_\theta^*(x_0|c) \propto p_{ref}(x_0|c) \exp\left(r(c,x_0)/\beta\right)$, one can eliminate $r$ and obtain a direct objective on $p_\theta$:
\begin{equation}
\begin{aligned}
    L = -\mathbb{E}_{c,x_0^{w/l}} \bigg[ \log \sigma \bigg(
    \beta \log \frac{p_\theta(x_0^w|c)}{p_\text{ref}(x_0^w|c)} - \beta \log \frac{p_\theta(x_0^l|c)}{p_\text{ref}(x_0^l|c)} \bigg) \bigg].
\end{aligned}
\end{equation}
This loss pushes generative distribution toward preferred outputs while keeping the learned policy not too far from the reference, avoiding potential reward hacking.

Extending DPO to diffusion models requires a tractable surrogate for the intractable parameterized distribution $p_\theta(x_0|c)$, as it requires marginalizing out all possible diffusion paths $(x_1, \dots, x_T)$ which lead to $x_0$.
To overcome this, Diffusion-DPO~\cite{wallace2024diffusionDPO} reformulates the objective on entire reverse trajectories $x_{0:T}$ rather than just the final samples $x_0$. This yields a new theoretical objective:
\begin{equation}
\begin{aligned}
L_\text{Diffusion-DPO}= -\mathbb{E}_{(x_0^w, x_0^l) \sim \mathcal{D}} \log \sigma \bigg( \
\beta \mathbb{E}_{\substack{x_{1:T}^w \sim p_\theta(x_{1:T}^w|x_0^w) \\ x_{1:T}^l \sim p_\theta(x_{1:T}^l|x_0^l)}}
\Big[ \\log \frac{p_\theta(x_{0:T}^w)}{p_\text{ref}(x_{0:T}^w)} - \log \frac{p_\theta(x_{0:T}^l)}{p_\text{ref}(x_{0:T}^l)} \Big] \bigg),
\end{aligned}
\end{equation}

Then it uses the ELBO together with an approximation that replaces the intractable reverse posterior by the forward noising process $q(x_{1:T}|x_0)$. After algebraic simplification and pushing expectations to a single timestep $t$, the training objective reduces to a preference-weighted denoising criterion. Writing $\epsilon_\theta$ for the model's noise prediction and $\epsilon_\text{ref}$ for the pretrained reference, the practical loss becomes
\begin{equation}
\label{eq:DiffusionDPO_final}
    L_\text{Diffusion-DPO}
=-\mathbb{E}_{x_0^w,x_0^l,t,x_t\sim q}
\log\sigma\Big(-\beta T\omega(\lambda_t)\big(\Delta^w-\Delta^l\big)\Big),
\end{equation}
with
$\Delta^\ast=\|\epsilon^\ast-\epsilon_\theta(x_t^\ast,t)\|_2^2-\|\epsilon^\ast-\epsilon_\text{ref}(x_t^\ast,t)\|_2^2$. 
$\lambda_t=\alpha_t^2/\sigma_t^2$ represents the signal-to-noise ratio, and $\omega(\lambda_t)$ denotes a weighting function, typically treated as a constant~\cite{ho2020ddpm, kingma2021variational}. The loss enables preference alignment for diffusion models without extra inference-time cost or unstable RL procedures.

\section{Method}
Based on above discussion, we find that human evaluation is inherently multi-dimensional, discrete, and non-equilibrium. Existing post-training frameworks~\cite{christiano2017deep, lee2023aligning, rafailov2023directDPO, wallace2024diffusionDPO} for generative models employ simplified signals, insufficient for capturing the intricate, fine-grained evaluation. 

To address this limitation, we propose a novel two-stage learning paradigm (\cref{fig:pipeline}), tailored for injecting and aligning with a complex, domain-specific criterion. First, we train the pretrained model to learn the evaluation attributes via Supervised Fine-Tuning (SFT) and thus form a domain-expert model. 
Second, by utilizing our proposed Complex Preference Optimization (CPO), we decouple the learning of positive and negative attributes in the alignment training.
Besides, we propose a new stabilization strategy.

\subsection{Domain-specific Knowledge Learning}
The objective of this stage is to develop an expert model that captures the correlation between training images and attributes defined in our domain-specific preference evaluation. 
The expert model is a text-to-image model parameterized by $\theta_1$ and initialized as the pre-trained $\theta_0$.
Specifically, our training data consists of tuples $(x_0, y, A_{pos}, A_{neg})$, where $x_0$ is the image, $y$ is the content description prompt, and $A_{pos}$ and $A_{neg}$ are the sets of positive and negative attribute labels, respectively.
The learning is conducted with Supervised Fine-Tuning (SFT) to minimize the denoising loss of \cref{eq: loss-diffusion}, 
where the condition $c$ is now a union of $y$, $A_{pos}$, and $A_{neg}$, and $\epsilon$ is the sampled ground-truth noise.

After fine-tuning, $\theta_1$ is aware of domain-specific knowledge. 
With prompt inputs augmented with $A_{pos}$ and $A_{neg}$ as auxiliary information during inference, $\theta_1$ generates images aligned with explicit textual attribute labels.
This model provides the foundation for the subsequent stage of preference learning.

\subsection{Complex Preference Optimization}

This stage performs implicit preference alignment. It trains the final model $\theta$ to generate images that conform to the domain-specific positive attributes $A_{pos}$ and eschew the negative $A_{neg}$.
Here $\theta$ is required to use only the content prompt $y$ as input.
The process decouples bipolar attributes in each sample by distilling knowledge from $\theta_1$ into $\theta$.

To achieve this, we introduce Complex Preference Optimization (CPO).
CPO is built on top of the Diffusion-DPO framework~\cite{wallace2024diffusionDPO}, an effective off-policy method to align models with human preferences. 
Instead of static, pre-defined pairs $(x^w, x^l)$ from a preference dataset, CPO leverages the SFT expert model $\theta_1$ as a dynamic reward oracle. At each denoising step $t$, a noisy sample $x_t$ is obtained based on $x_0$. For $x_t$, $\theta_1$ generates an ideal (winner) and non-ideal (loser) denoised prediction. 
These predictions are used to provide fine-grained guidance to $\theta$.

\textbf{Dynamic Process Reward Generation.}
Using the frozen expert model $\theta_1$, we construct two distinct conditional noise predictions at each timestep $t$, inspired by classifier-free guidance~\cite{ho2021classifierfree}. 

1. Winner Noise Prediction ($z^w$) represents the ideal denoising direction. It steers from a baseline of negative attributes toward the desired positive attributes. 
We define the positive conditioning $c_{pos} = (y, A_{pos})$ and the negative conditioning $c_{neg} = (A_{neg})$. The winner noise $z^w$ is:
\begin{equation}
    z^w(x_t, t) = (1-\omega_w)\epsilon_{\theta_1}(x_t, c_{neg}, t) + \omega_w \epsilon_{\theta_1}(x_t, c_{pos}, t).
\end{equation}

2. Loser Noise Prediction ($z^l$) represents the non-ideal direction. It steers from an unconditional baseline toward the explicitly negative attributes. We define $c_{all} = (y, A_{pos}, A_{neg})$ and $c_{null}=(\emptyset)$. The loser noise $z^l$ is:
\begin{equation}
    z^l(x_t, t) = (1-\omega_l) \epsilon_{\theta_1}(x_t, c_{null}, t) + \omega_l \epsilon_{\theta_1}(x_t, c_{all}, t).
\end{equation}
Here, $\omega_w$ and $\omega_l$ are hyperparameters both greater than 1.

\begin{figure}[t]
    \centering
    \includegraphics[width=1\linewidth]{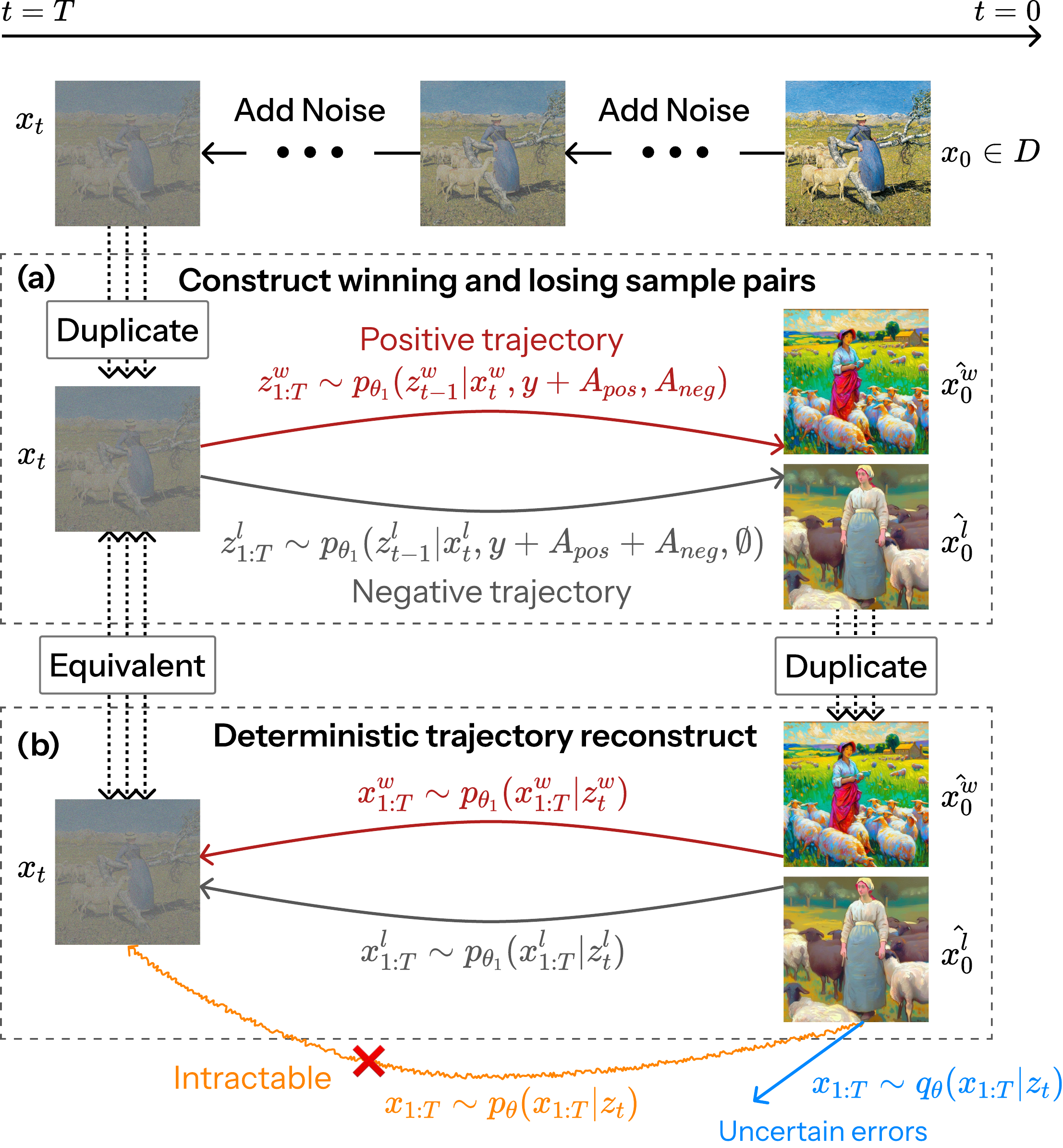} 
    \caption{Illustration of the CPO sampling trajectory. At each timestep $t$, CPO employs the expert model $\theta_1$ to provide deterministic positive and negative noise guidance, directing the trajectory toward virtual winning and losing samples, respectively. Owing to the determinism of the noise trajectory, the final sample $x_0$ can be precisely reconstructed back to $x_t$. Compared with original DPO, this design enables process-level guidance for model training rather than relying solely on the final endpoints, thereby making the training process more efficient.}
    \label{fig:trajectory}
\end{figure}

\textbf{CPO Objective}
The Diffusion-DPO approximates the intractable reverse process from a labeled sample to a posterior sample trajectory $p_{\theta}(x_{1:T}|x_{0})$ with the forward noising process $q(x_{1:T}|x_{0})$. 
This approximation, while necessary, inevitably introduces errors with  stochastic noise $\epsilon^w$ and $\epsilon^l$ drawn from $q$
as in \cref{eq:DiffusionDPO_final}.
Instead, in CPO we substitute the target noise to $z^w$ and $z^l$.

The rationale behind this substitution is stated as follows and is illustrated in \cref{fig:trajectory}.
First, suppose given a noisy $x_t$, a deterministic sampling process is conducted again with $z^w$ iteratively that biases toward positive away from negative attributes from $\theta_1$. The obtained $\hat{x}^w_0$ would have less negative evaluation than the original $x_0$. This is similar to $\hat{x}^l_0$ sampled from $z^l$. Therefore, we assume $\hat{x}^w_0$ is more preferrable than $\hat{x}^l_0$.
Second, given $\hat{x}^w_0$ and $\hat{x}^l_0$ as the winning and losing samples for DPO, the posterior trajectory is still intractable for $p_{\theta}(x_{1:T}|x_{0})$. Instead of using $q$, we propose to approximate $p_{\theta}(x_{1:T}|x_{0})$ with $p_{\theta_1}(x_{1:T}|z_t^w)$ for $\hat{x}^w_0$, which applies to $\hat{x}^l_0$ similarly. Conceptually, since the sampling processes are deterministic, the two approximated reverse trajectories overlap in $x_t$ again.
Third, for training efficiency, we focus on the training on $x_t$ only rather than all intermediate results on the trajectories. Since sampling iteratively with $z_w$ guides $x_t$ to the winning $\hat{x}^w_0$, $z_w$ is defined as the target of $\epsilon_\theta(\boldsymbol{x}_t,t)$ at step $t$. This is the same for $z_l$.
A detailed derivation with approximated KL divergence is in SM Sec. S7.

By incorporating these defined targets, we formulate the CPO loss $L_{CPO}$ to optimize $\theta$:
\begin{equation}
\label{eq: loss-cpo}
    \begin{aligned}
    L_{CPO}(\theta) =-\mathbb{E}_{\boldsymbol{x}_0\sim\mathcal{D},t\sim\mathcal{U}(0,T),\boldsymbol{z}_t^w,\boldsymbol{z}_t^l}
     \log\sigma(-\beta T\omega(\lambda_t)( \\
      \|\boldsymbol{z}^w-\boldsymbol{\epsilon}_\theta(\boldsymbol{x}_t,t)\|_2^2-\|\boldsymbol{z}^w-\boldsymbol{\epsilon}_\mathrm{ref}(\boldsymbol{x}_t,t)\|_2^2 \\
      -(\|\boldsymbol{z}^l-\boldsymbol{\epsilon}_\theta(\boldsymbol{x}_t,t)\|_2^2-\|\boldsymbol{z}^l-\boldsymbol{\epsilon}_\mathrm{ref}(\boldsymbol{x}_t,t)\|_2^2))) 
    \end{aligned}
\end{equation}
Note $\epsilon_{\theta}$ is only conditioned on the content prompt $y$. This objective explicitly encourages $\epsilon_{\theta}$ to minimize its error relative to the preferred noise $z^w$ and, conversely, to maximize its error relative to the dispreferred noise $z^l$. This mechanism allows the model to implicitly learn the positive attributes and unlearn the negative ones, decoupling attributes without requiring $A_{pos}$ or $A_{neg}$ at inference time.

\begin{table*}
  \caption{Quantitative results of SDXL- and FLUX-based methods on metrics evaluating attribute (\#A\_neg), quality (FID), and preference (the latter four metrics). $L_{\mathrm{CPO}}$ and $L_{\mathrm{CPO-S}}$ denote the training results without and with stabilization. $\ast$ denotes the comparison between our CPO, which does not require training a negative model, and NPO, which necessitates additional negative reward training.}
  \label{tab:quantitative}
  \centering
  \begin{tabular}{@{}lcccccc@{}}
    \toprule
    \footnotesize
    \textbf{Method} & \textbf{\#A\_neg (avg) $\downarrow$} & \textbf{FID} $\downarrow$ & \textbf{PickScore} $\uparrow$ & \textbf{HPSv2} $\uparrow$ & \textbf{ImageReward} $\uparrow$ & \textbf{Aesthetic} $\uparrow$ \\
    \midrule
    SDXL & 5.840 & 89.48 & 0.1963 & 0.2646 & 0.5180 & 6.210 \\
    SDXL-DPO & 5.790 & 93.12 & 0.2080 & 0.2906 & 0.9194 & 6.571 \\
    SDXL-SPO & 5.770 & 88.53 & 0.2081 & 0.2911 & 0.9200 & 6.577 \\
    SDXL-CPO ($L_{\mathrm{CPO}}$) & 5.210 & 88.07 & \textbf{0.2088} & 0.2918 & 0.9255 & \textbf{6.581} \\
    SDXL-CPO ($L_{\mathrm{CPO-S}}$) $\ast$ & \textbf{5.180} & \textbf{87.37} & 0.2083 & \textbf{0.3039} & \textbf{0.9312} & \textbf{6.581} \\
    \midrule
    SDXL+NPO $\ast$ & 5.210 & 84.88 & \textbf{0.2120} & 0.2786 & 0.8729 & 6.541 \\
    SDXL-DPO+NPO & 5.630 & 86.93 & 0.2084 & 0.2960 & \textbf{0.9992} & 6.539 \\
    SDXL-CPO+NPO & \textbf{5.070} & \textbf{79.13} & 0.2118 & \textbf{0.2989} & 0.9784 & \textbf{6.591} \\
    \midrule
    FLUX & 5.120 & \textbf{95.69} & 0.2005 & 0.2853 & 0.8696 & 6.460 \\
    FLUX-DPO & 4.400 & 104.79 & \textbf{0.2113} & 0.3210 & 1.1516 & 6.864 \\
    FLUX-CPO & \textbf{3.780} & 104.71 & \textbf{0.2113} & \textbf{0.3212} & \textbf{1.1526} & \textbf{6.865} \\
        
    \bottomrule
  \end{tabular}
\end{table*}

\subsection{Stabilization of the Optimization}

Empirically, we observed that training with the standard DPO-style objective suffers from instabilities. We attribute this to the imbalance of winning and losing parts in optimization. The losing term $-\|z^l-\epsilon_\theta(x_t^l,t)\|_2^2$ is innately concave, and the resulting gradient norm grows as the training proceeds.
However, the winning term $\|z^w-\epsilon_\theta(x_t^w,t)\|_2^2$ is convex instead, and its gradient norm shrinks.
Therefore, the gradient norm of the losing term grows disproportionately compared to the winning. Such phenomenon also applies to other methods based on DiffusionDPO.

To address this, we stabilize our CPO objective by transforming the original loss with another term. The aim is to ensure the gradient of the losing term is equal to that of the winning term.
This ensures that the optimization landscape remains stable and that the gradients from the winner and loser terms are balanced. Specifically, we define
\begin{equation}
    z^{l-tgt}=\epsilon_\theta(x_t,t) + \frac{\epsilon_\theta(x_t,t) - z^l}{\| \epsilon_\theta(x_t,t) - z^l \|} \|\epsilon_\theta(x_t,t) - z^w\|.
\end{equation}
Our stabilized objective, $L_{CPO-S}$, is formulated as:
\begin{equation}
\label{eq: loss-cpo-s}
    \begin{aligned}
    &L_{CPO-S}(\theta) =-\mathbb{E}_{\boldsymbol{x}_0\sim\mathcal{D},t\sim\mathcal{U}(0,T),\boldsymbol{z}_t^w,\boldsymbol{z}_t^l}
    \log\sigma( \\
    &-\beta T\omega(\lambda_t)
      (\|\boldsymbol{z}^w-\boldsymbol{\epsilon}_\theta(\boldsymbol{x}_t,t)\|_2^2-\|\boldsymbol{z}^w-\boldsymbol{\epsilon}_\mathrm{ref}(\boldsymbol{x}_t,t)\|_2^2 \\
      &+(\|\boldsymbol{z}^{l-tgt}-\boldsymbol{\epsilon}_\theta(\boldsymbol{x}_t,t)\|_2^2-\|\boldsymbol{z}^{l-tgt}-\boldsymbol{\epsilon}_\mathrm{ref}(\boldsymbol{x}_t,t)\|_2^2))).
    \end{aligned}
\end{equation}
In the implementation, we apply a stop-gradient (detachment) operation to $z^{l-tgt}$.
In this case, the direction of the gradient backward to $\boldsymbol{\epsilon}_\theta(\boldsymbol{x}_t^l,t)$ is the same as the original loss but the norm is restricted to $\|\boldsymbol{z}^w-\boldsymbol{\epsilon}_\theta(\boldsymbol{x}_t^w,t)\|$.
A more detailed derivation and analysis can be found in Sec. S5 of the Supplementary Material.

This stabilization ensures that the loser term's contribution to the gradient is balanced with that of the winner term.
Theoretically, a surrogate convex term is used to substitute the original concave term, leading to significantly more robust convergence as shown in empirical results.

\begin{figure*}[t]
    \centering
    \includegraphics[width=0.9\linewidth]{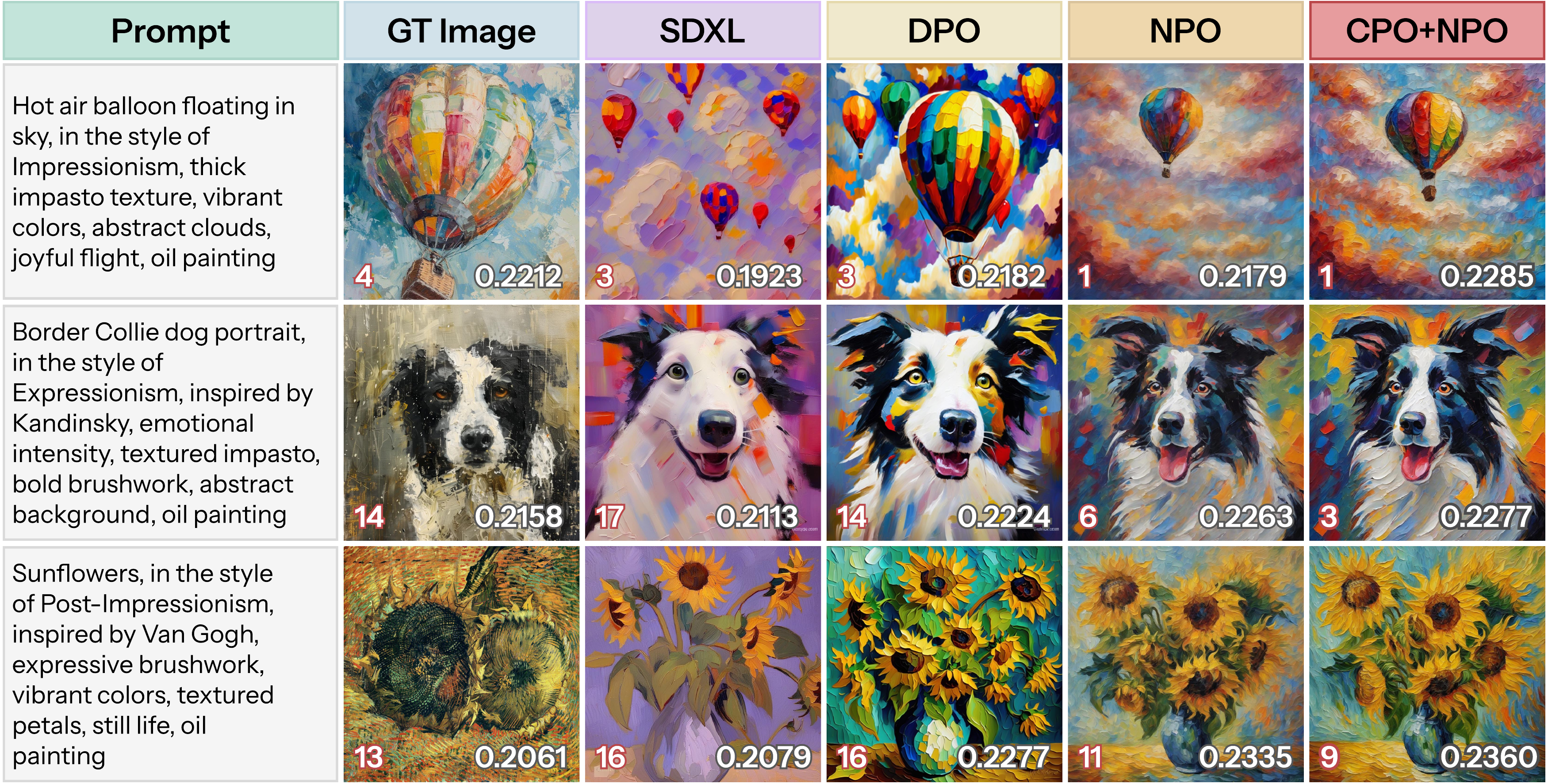} 
    \caption{Visual comparison of different baselines and our CPO. \#A\_neg ($\downarrow$) and PickScore ($\uparrow$) are annotated in the lower-left and lower-right corners of each image, respectively. CPO outperforms all baselines in both negative-attribute avoidance and preference scoring.}
    \label{fig:qualitative}
\end{figure*}

\section{Experiment}
\subsection{Dataset and Implement}
We collect 10,277 diverse publicly available paintings with automated filtering and manual inspection. The dataset is randomly split into 8,221 (80\%) / 1,028 (10\%) / 1,028 (10\%) images for training, validation, and testing. With this dataset, we train our models in two stages. In Stage 1, we perform supervised fine-tuning on the base model using LoRA~\cite{hu2022lora} over the full dataset. Each training instance concatenates the base prompt with its positive and negative labels into a single textual input. We use a LoRA rank of 16, a learning rate of 1e-4, and train for two epochs. In Stage 2, for SDXL, we follow the Diffusion-DPO training configuration~\cite{wallace2024diffusionDPO} to ensure fair comparison, training for 8,221 steps (one epoch). For FLUX, we apply LoRA-based post-training with a reduced rank of 8, keeping the 1e-4 learning rate and setting the LoRA scaling factor $\beta$ to 0.1. All experiments are conducted on a single NVIDIA H800 GPU.

\subsection{Evaluation and Baselines}
We introduce \#A\_neg as a new metric, quantifying the presence of negative attributes identified by our domain-expert agent (\cref{sec:Domain-specific Fine-grained Evaluation}) and averaged over 300 images. We also conduct evaluation on existing metrics for general image quality, aesthetics, and human preference, including FID~\cite{NEURIPS2023_dd83eada_FID}, PickScore~\cite{Yuval2023Pick-a-Pic}, HPSv2~\cite{wu2023hpsv2}, ImageReward~\cite{xu2024ImageReward}, and Aesthetic Score~\cite{Schuhmann2022LAION_5B}. We compare our CPO against baseline methods on both SDXL- and FLUX- based models. Baselines include the fine-tuned SDXL~\cite{podell2023sdxlimprovinglatentdiffusion} and FLUX~\cite{blackforest2024flux} as well as Diffusion-DPO~\cite{wang2025diffusionnpo}, SPO~\cite{liang2025aestheticSPO}, and their NPO-augmented~\cite{lu2025inpo} variants. We also report our non-stabilized objective, SDXL-CPO ($L_{CPO}$), as an ablation result.

\subsection{Quantitative Result}
As shown in \cref{tab:quantitative}, our CPO demonstrates clear superiority. On the primary SDXL group, our stabilized method SDXL-CPO ($L_{\mathrm{CPO-S}}$) excels in avoiding negative attributes, significantly reducing \#A\_neg to 5.180. Critically, this reduction does not compromise quality: our method simultaneously secures the best FID (87.37) and joint-highest preference scores. We report the number of negative rather than positive attributes. This is because human expertise is inherently non-equilibrium(\cref{sec:Domain-specific Fine-grained Evaluation}), meaning images are assessed under varying criteria. Consequently, a high \#A\_pos does not necessarily indicate better image quality. 
In contrast, the presence of negative attributes is \textit{consistently} undesirable, 
making \#A\_neg a more reasonable evaluation metric.

Compared with CPO, NPO~\cite{lu2025inpo} requires additional training of a negative reward model.
NPO underperforms CPO on most metrics (see the two rows marked with $\ast$ in \cref{tab:quantitative}). When all methods are further trained with NPO, our CPO still demonstrates superior overall performance, showing only lower scores on PickScore and ImageReward. 

CPO generalizes robustly to other architectures. 
On the FLUX-based model, FLUX-CPO achieves a \#A\_neg score of 3.780, a dramatic improvement over both the FLUX baseline (5.120) and FLUX-DPO (4.400). It also achieves the best results in preference scores. The increase in FID is unavoidable following fine-tuning, which is also reported in existing research~\cite{shen2024sgadapter}~\cite{he2025robust}~\cite{wang2024reward}~\cite{li2024laion}, and this effect is particularly pronounced within FLUX~\cite{Chen_2025_SANA-Sprint}.




\subsection{Qualitative Result}
\cref{fig:qualitative} illustrates the visual performance of our CPO compared to baseline models. Each row shows an input prompt and the corresponding generated images. Images generated by CPO exhibit the fewest negative attributes (marked in red), which is also evident from the last column---our results consistently demonstrate superior composition, color harmony, light and shadow, and brushstroke quality. CPO further tends to achieve higher preference scores, for which we report the PickScore (marked in grey) as an instance.

\subsection{Ablation Study}
We conduct ablations to validate our framework, including the necessity of our fine-grained, attribute-decoupled reward design, the impact of the training data volume, and the effectiveness of the stabilization strategy.

\textbf{Impact of Reward Granularity.} As shown in \cref{tab:ablation-diff-reward}, we compare our 7-dimensional complex reward against two coarser-grained reward structures. Scalar denotes normalizing and averaging the 7 dimensions into a single score. And binary denotes simplifying each of the 7 dimensions into a ``winning"/``losing" label. The results clearly indicate that model performance scales directly with the granularity of the feedback signal. Our complex reward performs the best across all metrics. This finding compellingly demonstrates that our proposed complex preference optimization is critical for achieving desired alignment.

\textbf{Impact of Data Proportion.} In \cref{tab:ablation-data-prop}, we analyze the effect of data volume by training with varying proportions of our attribute-decoupled dataset. The results show a significant improvement as the dataset size increases.

\begin{figure}[t]
    \centering
    \includegraphics[width=1\linewidth]{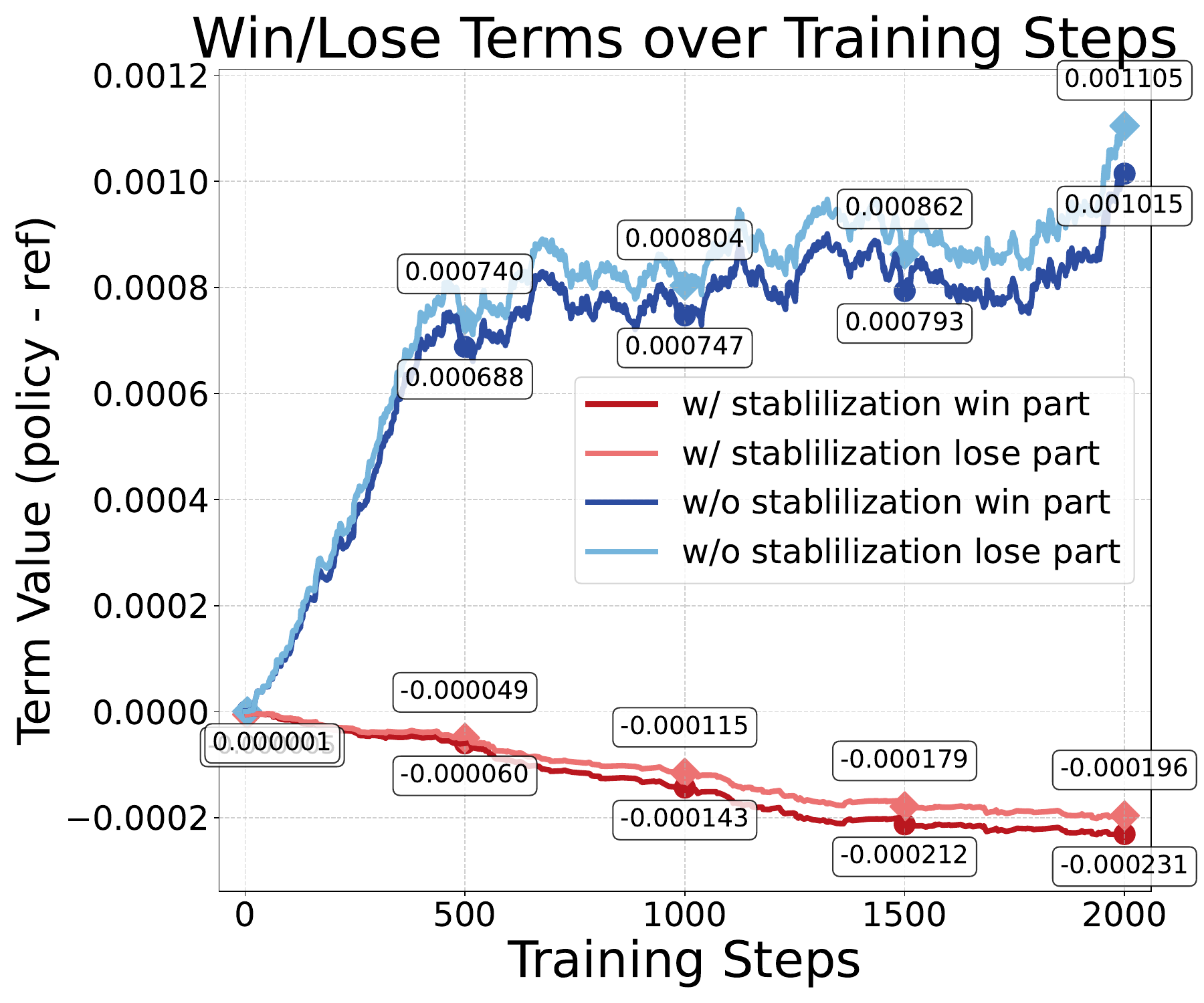} 
    \caption{Curves of the win and lose parts of the loss function over training steps. The configuration with stabilization demonstrates significantly greater stability compared to the one without.}
    \label{fig:loss-curve}
\end{figure}

\textbf{Effectiveness of stabilization strategy.} \cref{fig:loss-curve} plots the values of the winning and losing parts in \cref{eq: loss-cpo} ($L_{\mathrm{CPO}}$) over training steps. The winning part is $\|\boldsymbol{z}^w-\boldsymbol{\epsilon}_\theta(\boldsymbol{x}_t,t)\|_2^2-\|\boldsymbol{z}^w-\boldsymbol{\epsilon}_\mathrm{ref}(\boldsymbol{x}_t,t)\|_2^2$ and the losing is similar.
Given our stabilization, both parts exhibit a markedly smoother and more stable decrease, whereas the loss without stabilization undergoes substantial oscillations. 
Notice that the winning part is expected to be minimized while the losing is maximized.
The joint change of both loss term is known as gradient entanglement~\cite{gradient_entanglement} and widely observed. 
Here, the original loss emphasizes more on the unlearning the losing but fail to optimize the winning part. Our stabilization allows the optimization to emphsize on learning the positive attrbutes over unlearning negative ones.
The superior performance of our $L_{\mathrm{CPO-S}}$ over $L_{\mathrm{CPO}}$ (refer to \cref{tab:quantitative}) also confirms the efficacy of our stabilization strategy.

\begin{table}
  \caption{Comparison of different reward designs. Scalar and Binary denote scalar score–based and binary preference–based optimization, respectively, while Complex represents our fine-grained, attribute-decoupled preference optimization.}
  \label{tab:ablation-diff-reward}
  \centering
  \footnotesize
  \begin{tabular}{@{}lcccccc@{}}
    \toprule
    
    \textbf{Reward} & \textbf{\#AN}$^{\downarrow}$ & \textbf{FID}$^{\downarrow}$ & \textbf{PS}$^{\uparrow}$ & \textbf{HPS}$^{\uparrow}$ & \textbf{IR}$^{\uparrow}$ & \textbf{LA}$^{\uparrow}$ \\
    \midrule
    Scalar & 5.840 & 91.99 & 0.1959 & 0.2649 & 0.5194 & 6.239 \\
    Binary & 5.270 & 87.43 & 0.2080 & 0.2921 & 0.9296 & 6.577 \\
    Complex & \textbf{5.180} & \textbf{87.37} & \textbf{0.2083} & \textbf{0.3039} & \textbf{0.9312} & \textbf{6.581} \\
    
    \bottomrule
  \end{tabular}
\end{table}

\begin{table}
  \caption{Ablation study under different proportions (Prop.) of attribute-decoupled training data. AN, PS, IR, and LA denote \#A\_neg, PickScore, ImageReward, and LAION-Aesthetic.}
  \label{tab:ablation-data-prop}
  \centering
  \footnotesize
  \begin{tabular}{@{}lcccccc@{}}
    \toprule
    
    \textbf{Prop.} & \textbf{\#AN}$^{\downarrow}$ & \textbf{FID}$^{\downarrow}$ & \textbf{PS}$^{\uparrow}$ & \textbf{HPS}$^{\uparrow}$ & \textbf{IR}$^{\uparrow}$ & \textbf{LA}$^{\uparrow}$ \\
    \midrule
    10\%   & 5.770 & 89.39 & 0.2066 & 0.2905 & 0.9122 & 6.562 \\
    20\%   & 5.750 & 89.37 & 0.2073 & 0.2914 & 0.9266 & 6.566 \\
    50\%   & 5.530 & \textbf{86.61} & 0.2074 & 0.2917 & 0.9311 & 6.566 \\
    100\%  & \textbf{5.180} & 87.37 & \textbf{0.2083} & \textbf{0.3039} & \textbf{0.9312} & \textbf{6.581} \\
    
    \bottomrule
  \end{tabular}
\end{table}

\section{Conclusion}
We aim to address the reliance on simplified feedback in preference alignment, introducing a hierarchical, fine-grained evaluation criterion with positive and negative attributes. Based on this, we propose a two-stage alignment with a stabilization strategy to learn complex expertise. Experiments demonstrate CPO outperforms existing baselines. 
{
    \small
    \bibliographystyle{ieeenat_fullname}
    \bibliography{main}
}

\newpage 
\begin{center}
    \section*{Supplementary Material}
\end{center}
\renewcommand{\thesection}{S\arabic{section}}
\renewcommand{\thetable}{S\arabic{table}}
\renewcommand{\thefigure}{S\arabic{figure}}
\renewcommand{\theequation}{S\arabic{equation}}

\setcounter{section}{0}
\section{Description of the Fine-grained Hierarchical Evaluation}
\begin{figure*}[t]
    \centering
    \includegraphics[width=1\linewidth]{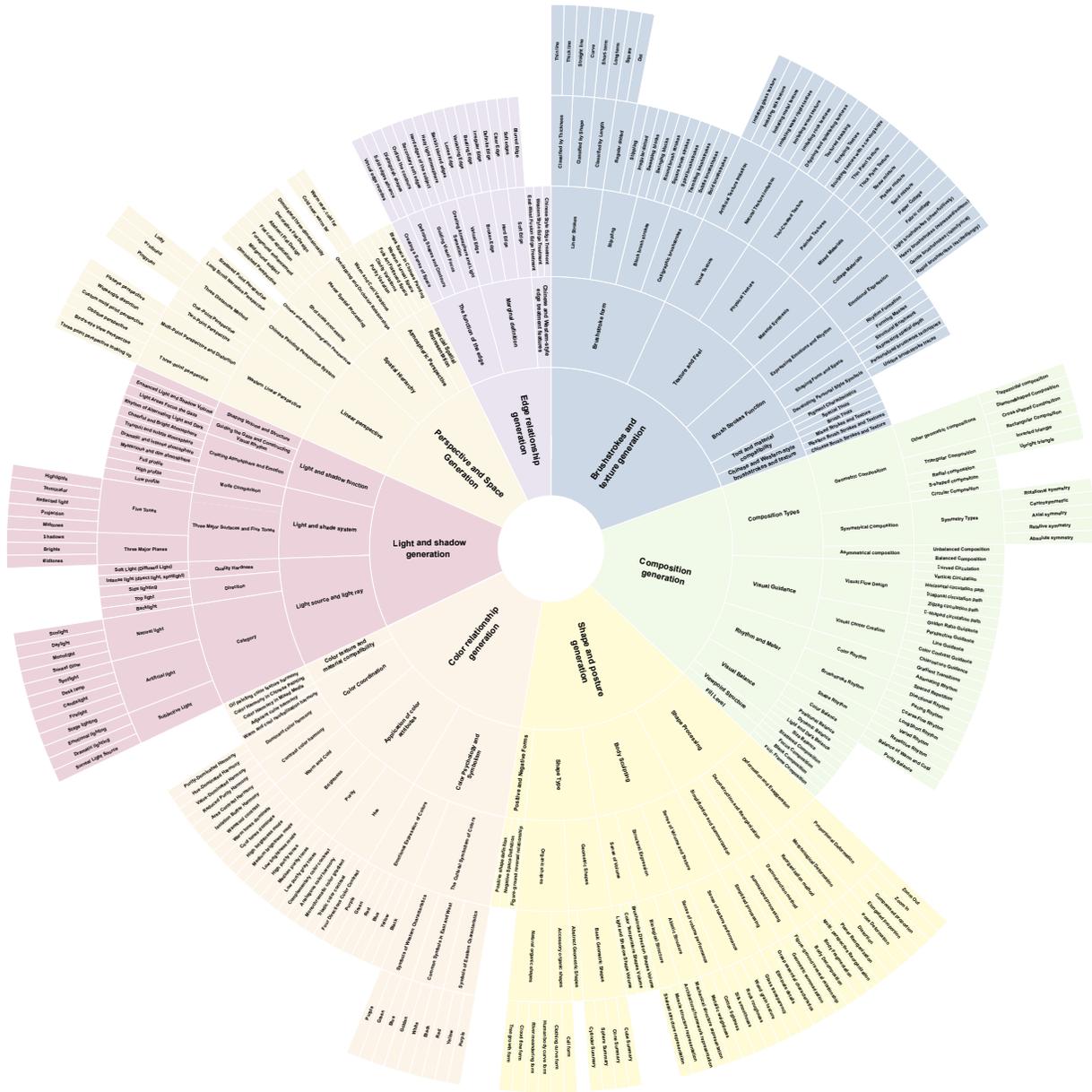} 
    \caption{Illustration of the domain-specific fine-grained evaluation framework. Best viewed magnified on screen.}
    \label{fig:sunfigure} 
\end{figure*}

As referenced in Section 3 of the main paper, our domain-specific fine-grained evaluation system operates based on a 5-level knowledge hierarchy comprising 7 root dimensions and a comprehensive set of 246 attribute pairs, assessing multiple aspects such as Composition, Color Relationships, and Brushstrokes \& Texture. The complete structure is visualized in \cref{fig:sunfigure}. This structure underpins that human evaluation is inherently multi-dimensional, discrete, and non-equilibrium. The fine-grained evaluation system serves as the foundational knowledge base for our Complex Preference Optimization (CPO) framework, addressing the limitations of coarse, simplified feedback signals used in prevailing alignment methods.

This hierarchical paradigm provides a critical advantage over standard text-based evaluation, which relies on monolithic image-level reward signals. The core difference lies in the granularity and bidirectional control. Our system explicitly encodes knowledge spanning seven root dimensions and five hierarchical levels, encompassing specialized sub-dimensions such as Visual Guidance under Composition and Light Aspect/Quality under Light and Shadow. This fine-grained evaluation scheme enhances the model’s capacity to perceive and learn domain-specific knowledge. Most importantly, it enables decoupled supervision by providing separate positive ($A_{pos}$) and negative ($A_{neg}$) attribute sets for the same image. This design is essential because negative attributes often coexist with positive ones---an image is rarely uniformly good or bad across all aspects. Such granularity and explicit bidirectional control allow CPO to learn complex expert criteria, delivering precise, attribute-level guidance that cannot be achieved by monolithic rewards derived from simple text prompts.

\section{Description of Complex Preference Learning Tasks}
\begin{figure}[t]
    \centering
    \includegraphics[width=1\linewidth]{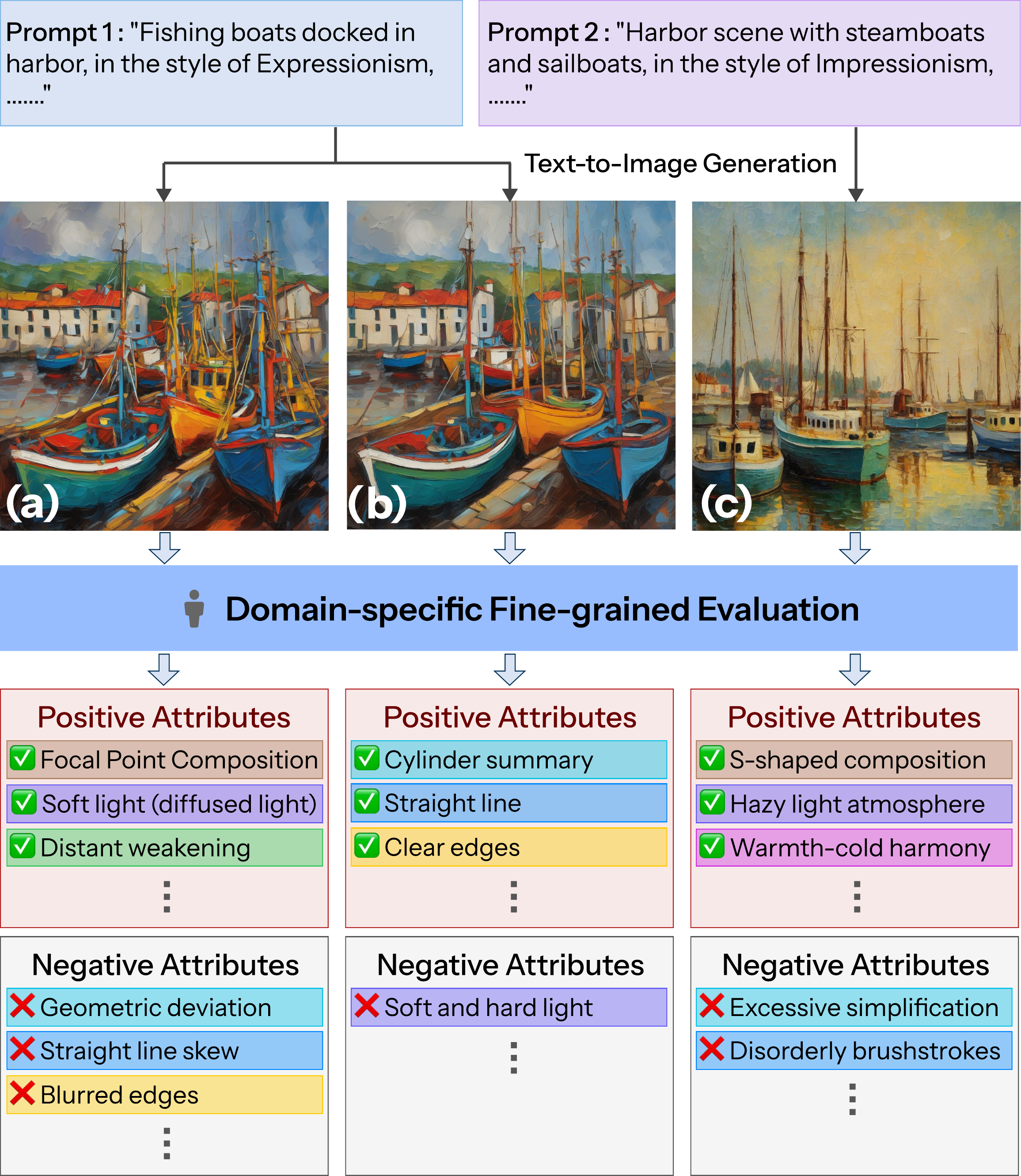} 
    \caption{Description of tasks targeted by CPO. Image (a) and (b) are generated from the same prompt, yet each exhibits its own strengths and weaknesses; thus, it is inappropriate to generalize that either image is universally superior. Image (c), generated from a different prompt, should be evaluated using criteria distinct from those applied to (a) and (b).}
    \label{fig:SM-task}
\end{figure}

Human cognitive evaluation is inherently not a highly regularized process. By collaborating with domain experts to construct the evaluation criteria, we observe that human evaluation is inherently multidimensional, discrete, and non-equilibrium, which is consistent with findings reported in prior research~\cite{freedman2001categorical, treisman1980feature, leder2004model}. For example, as illustrated in \cref{fig:SM-task}, an expert evaluating two paintings images may identify one (\cref{fig:SM-task}(a)) as having positive ``Focal Point Composition" but negtive ``Blurred edges", while another (\cref{fig:SM-task}(b)) exhibits ``Clear edges" yet suffers from ``Soft and hard light". Besides, the evaluation labels of \cref{fig:SM-task}(c) vary with changes in content and style, reflecting the non-equilibrium of human evaluation. Furthermore, complex, discrepancy and non-equilibrium mean that multi-dimensional reward functions should not be used for simple scoring, and the multi-dimensional nature is not suited for directly assigning a single notion of superiority or inferiority. Therefore, it is essential to develop a new paradigm aligning with human evaluation and to formulate corresponding algorithms.


\section{User Study}
\label{sec:user_study}
\begin{figure*}[t] 
    \centering
    
    
    \begin{minipage}{1.0\linewidth}
        \centering
        \footnotesize \textbf{
        \makebox[0.24\linewidth]{SDXL-DPO+NPO} \hfill
        \makebox[0.24\linewidth]{SDXL-CPO+NPO} \hfill
        \makebox[0.24\linewidth]{FLUX-DPO} \hfill
        \makebox[0.24\linewidth]{FLUX-CPO}
        }
    \end{minipage}
    \vspace{0.2em}
    
    \includegraphics[width=0.24\linewidth]{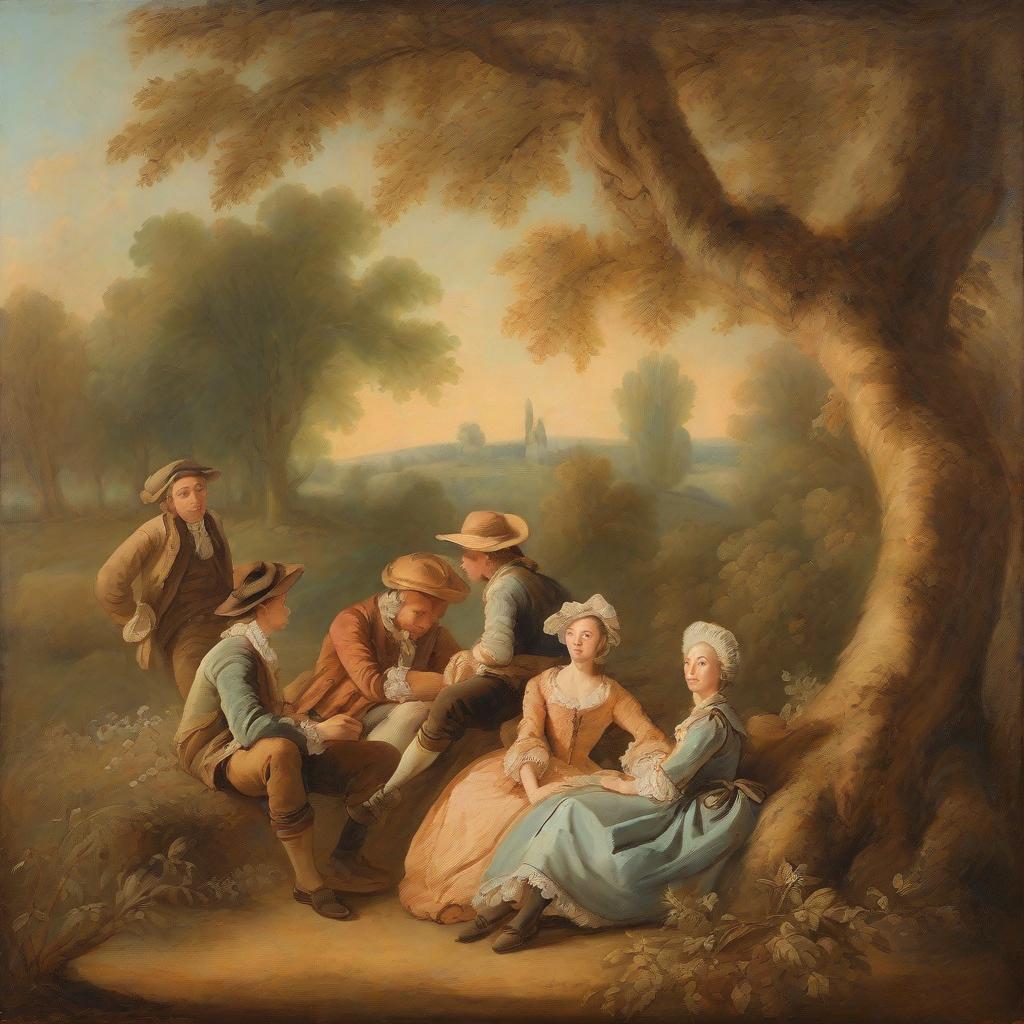} \hfill
    \includegraphics[width=0.24\linewidth]{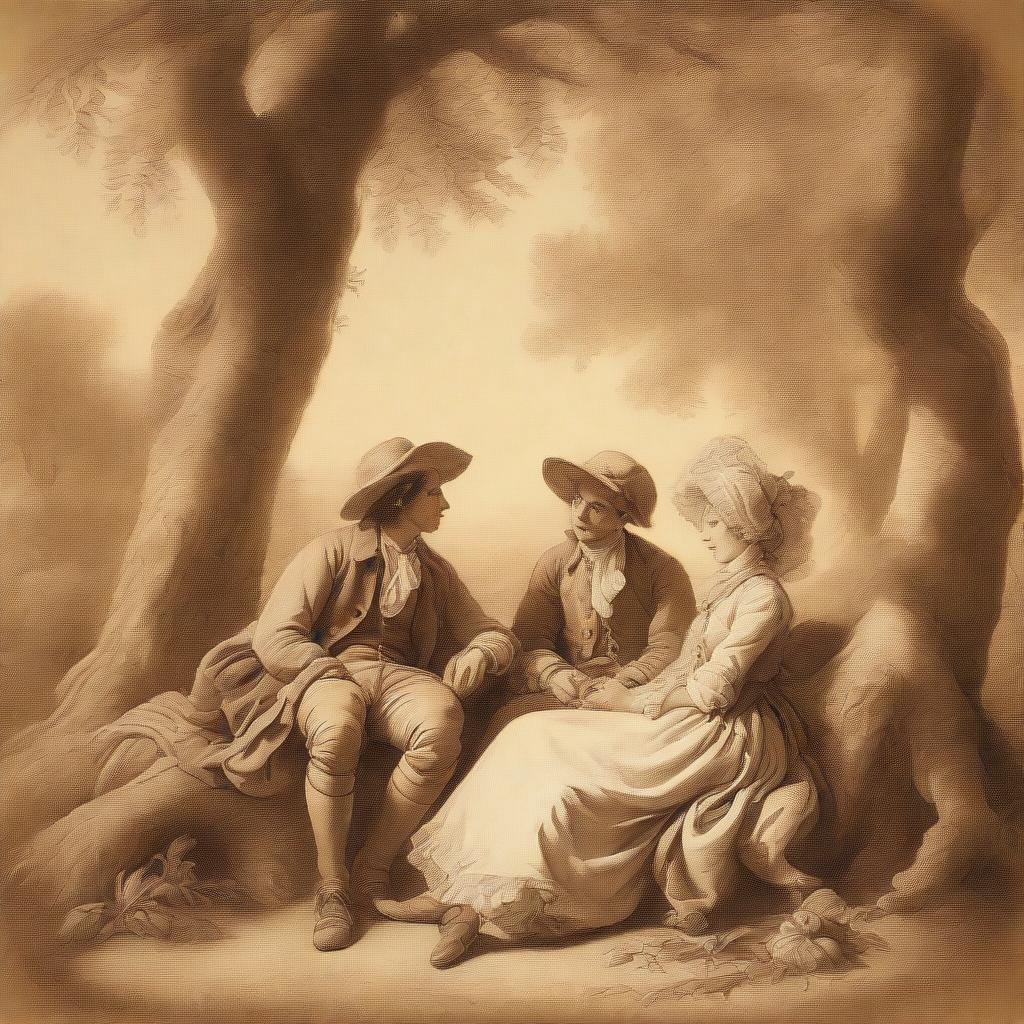} \hfill
    \includegraphics[width=0.24\linewidth]{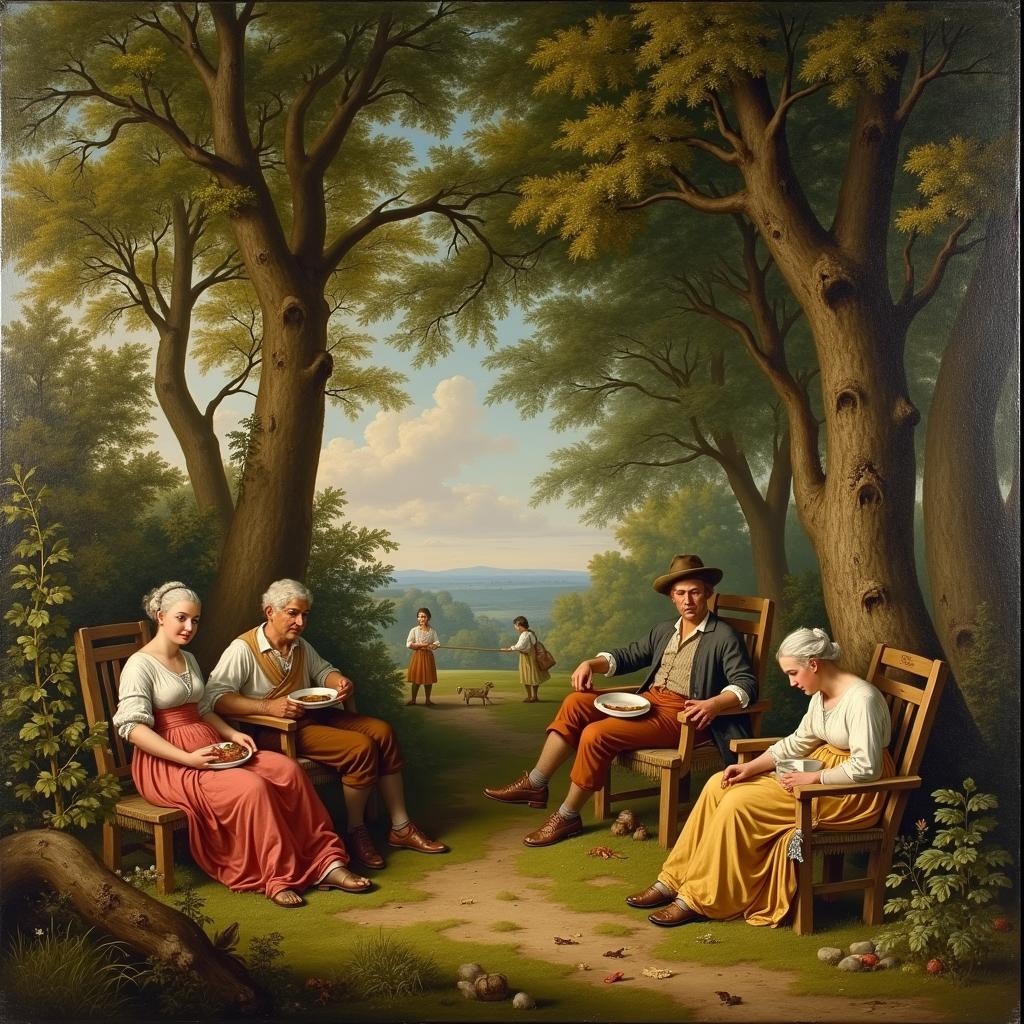} \hfill
    \includegraphics[width=0.24\linewidth]{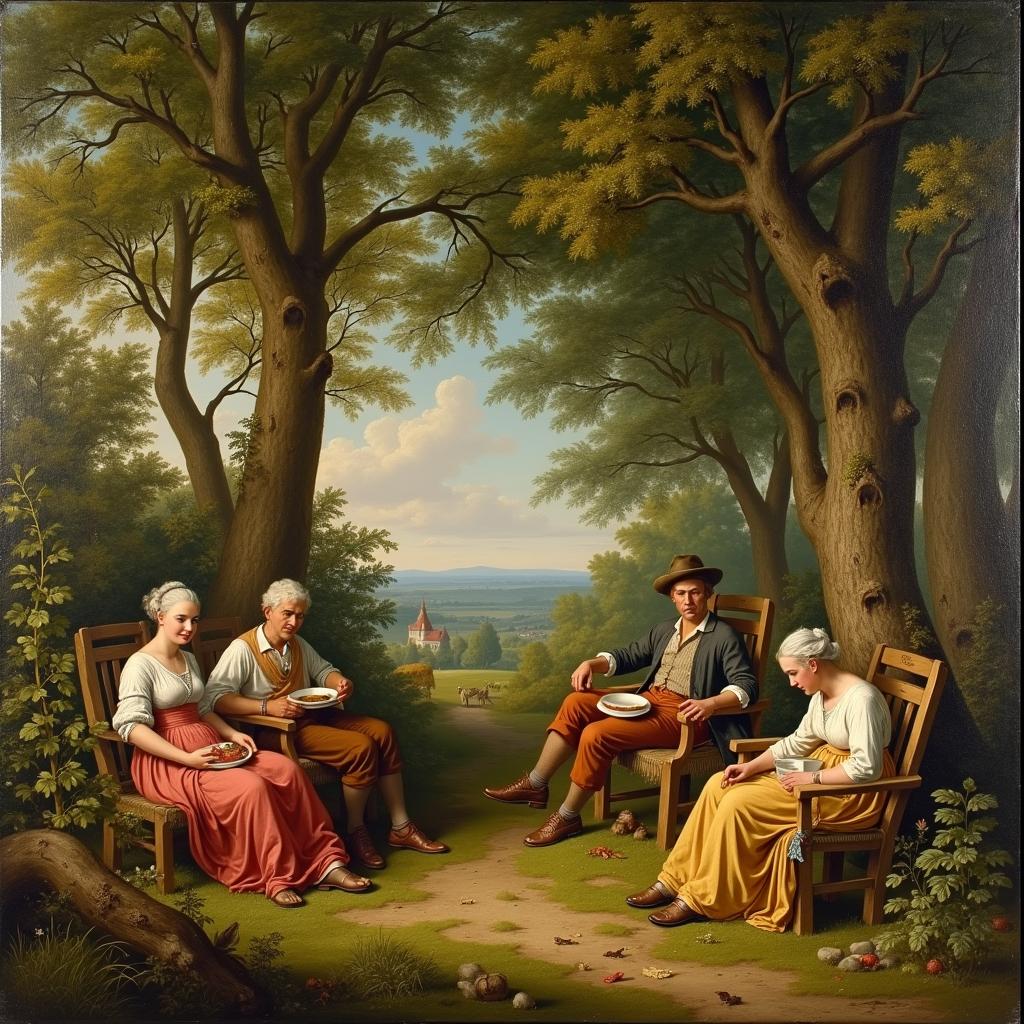}
    
    \vspace{0.3em}
    \begin{minipage}{0.98\linewidth}
        \small 
        Still life with bottle and fruit, in the style of Expressionism, inspired by Karl Schmidt-Rottluff, bold brushwork, vibrant color, simplified form, textured surface, oil painting
    \end{minipage}
    \vspace{0.8em} 
    
      \includegraphics[width=0.24\linewidth]{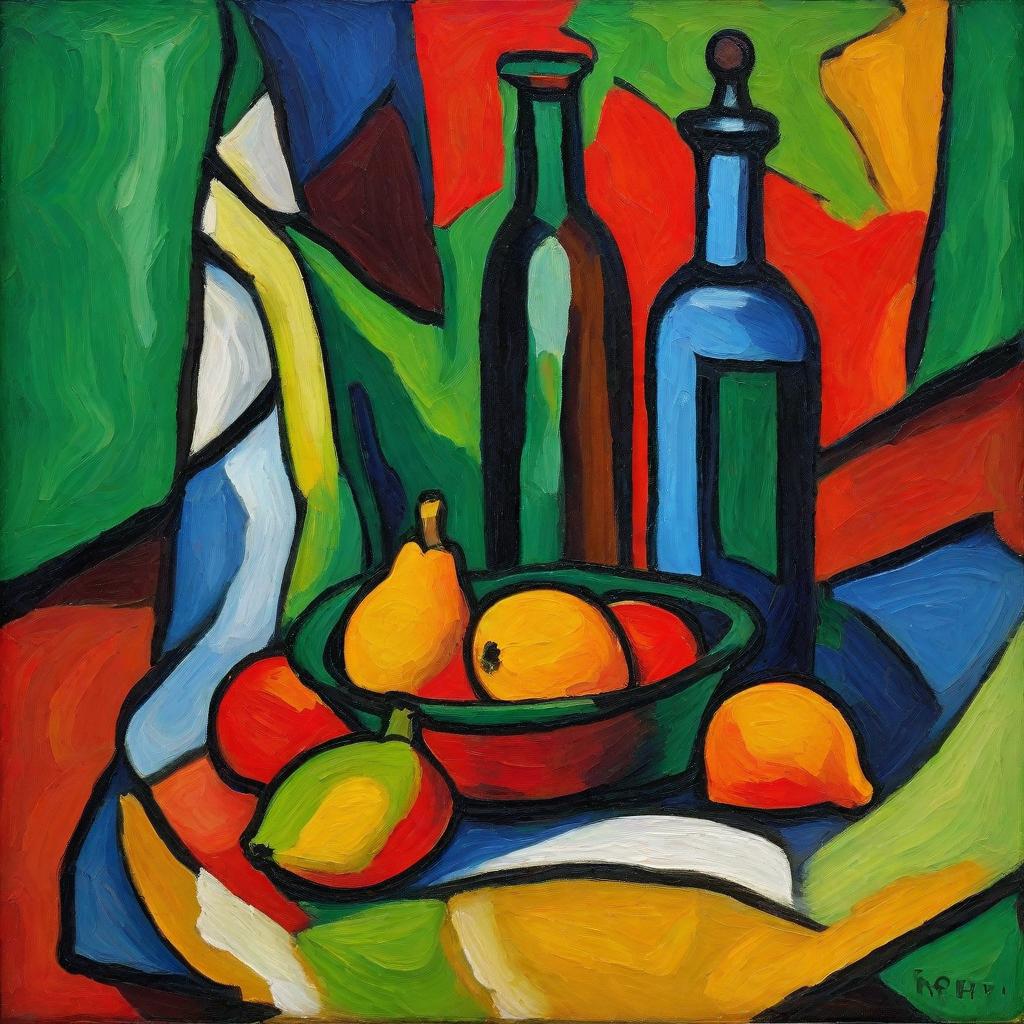} \hfill
    \includegraphics[width=0.24\linewidth]{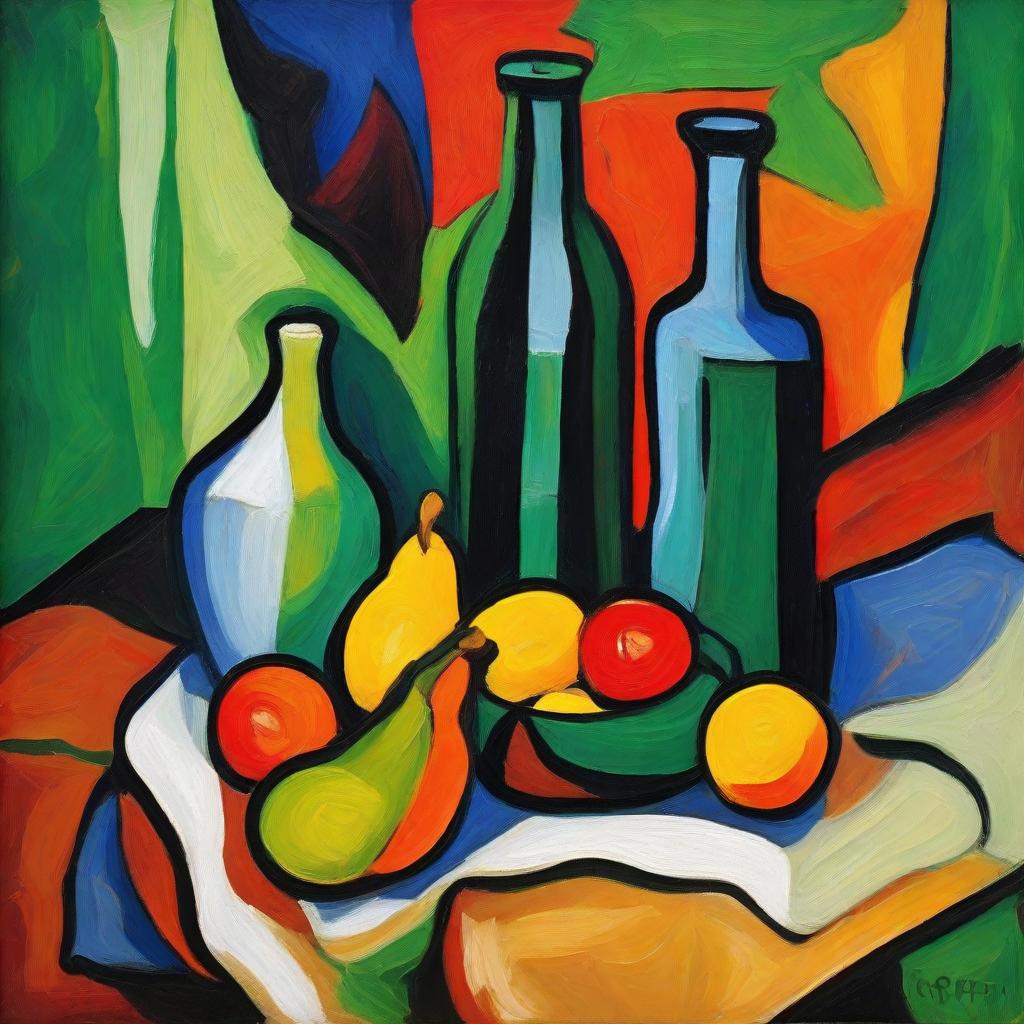} \hfill
    \includegraphics[width=0.24\linewidth]{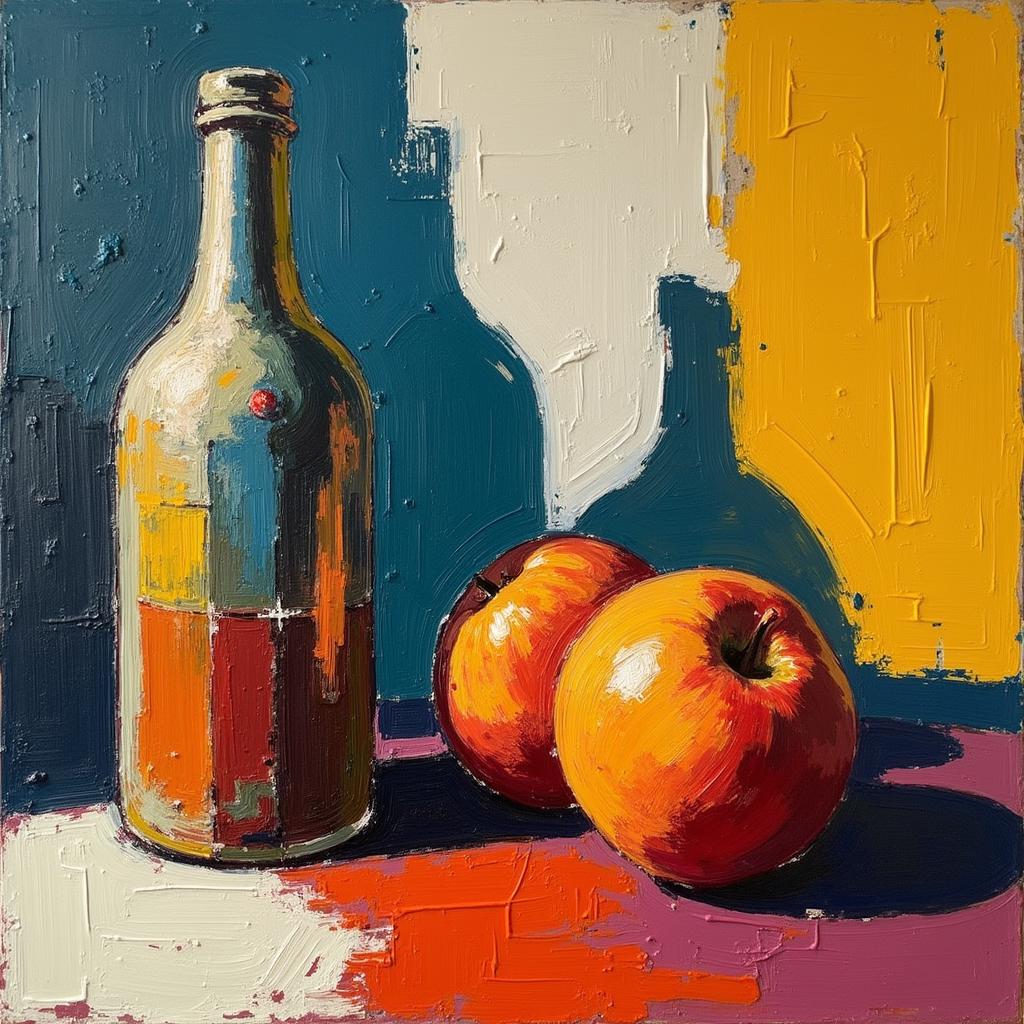} \hfill
    \includegraphics[width=0.24\linewidth]{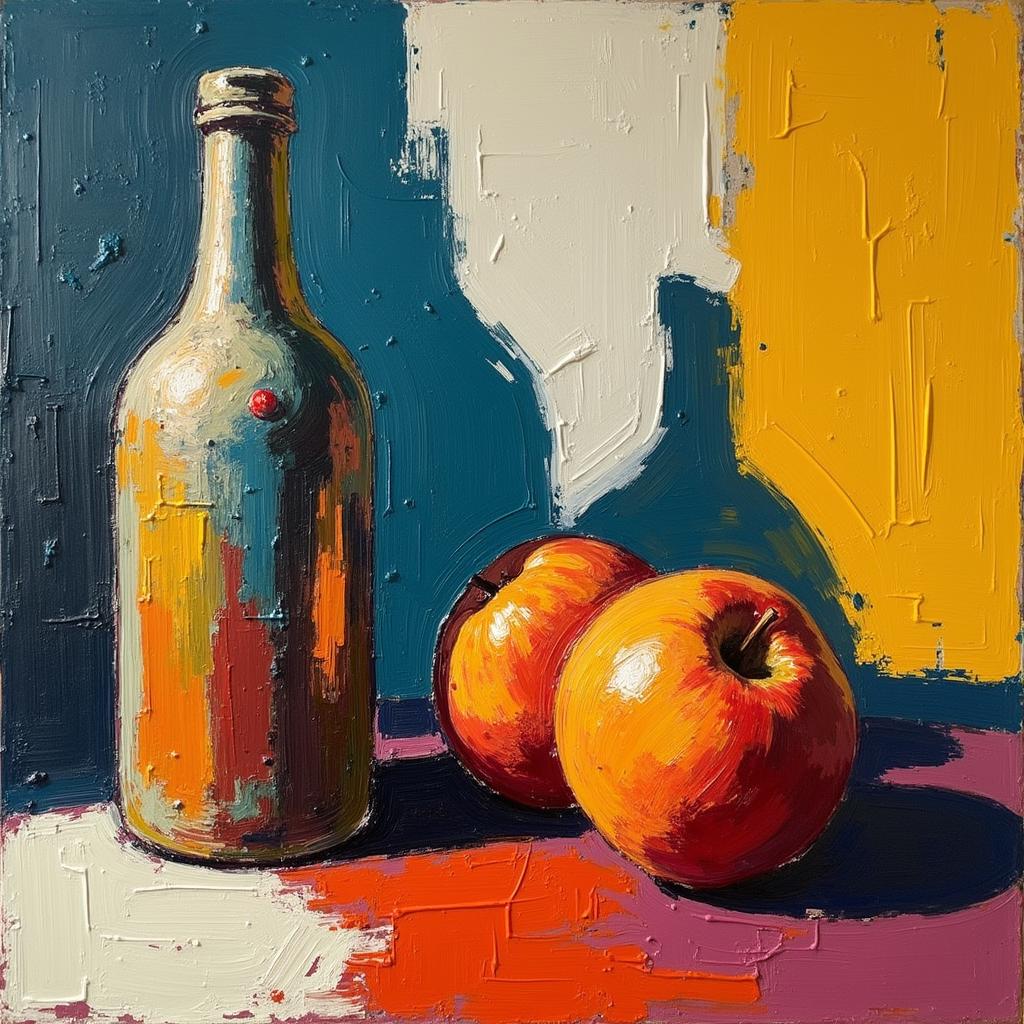}
    
    \vspace{0.3em}
    \begin{minipage}{0.98\linewidth}
        \small
        Peasants resting under trees, in the style of Rococo, inspired by Watteau, pastoral, rustic, rural life, outdoor gathering, warm light, oil painting
    \end{minipage}
    
    \vspace{0.5em} 
    
    \caption*{\textit{Please use the following 7 dimensions as criteria to conduct pairwise comparisons for the image pairs in Group G1 and Group G2, respectively. For each dimension, select the image that performs better: Brushwork and Texture Generation, Edge Relationship Generation, Composition Generation, Light and Shadow Generation, Color Relationship Generation, Perspective and Space Generation, and Shape and Form Generation.}}
    
    \vspace{0.2em}
    
    \begin{tabular}{@{}l cc cc@{}}
        \toprule
        Group & \multicolumn{2}{c}{G1} & \multicolumn{2}{c}{G2} \\
        \cmidrule(lr){2-3} \cmidrule(l){4-5}
        Model & SDXL-DPO+NPO & SDXL-CPO+NPO & FLUX-DPO & FLUX-CPO \\
        \midrule
        User Preference & 36.5\% & 63.5\% & 15.9\% & 84.1\% \\
        \bottomrule
    \end{tabular}
    
    \vspace{0.5em}
    
    \caption{The result of user study. \textbf{Top:} Qualitative comparison of images generated by different methods using the same prompt. \textbf{Bottom:} Quantitative results from the user study showing preference rates for our CPO methods against DPO baselines across two base models (SDXL and FLUX).}
    \label{fig:user_study}
    
\end{figure*}

To validate whether our proposed CPO (Complex Preference Optimization) method can generate images more aligned with complex human perception than baseline methods (Diffusion-DPO~\cite{wallace2024diffusionDPO}, Diffusion-NPO~\cite{wang2025diffusionnpo}), we design and execute a user study. The core purpose of this study is to compare the subjective visual quality of images generated by different models from a professional perspective.

We first randomly sample 150 prompts from the test set. Subsequently, we use these prompts and feed them separately into the following four trained models: SDXL-DPO+NPO, SDXL-CPO+NPO, FLUX-DPO, FLUX-CPO, generating a total of 600 images for evaluation.

In each trial, participants  observe two groups (G1, G2) of images generated from the same prompt. Group G1 (SDXL Base) contain an image generated by SDXL-DPO+NPO and an image generated by SDXL-CPO+NPO. Group G2 (FLUX Base) contain an image generated by FLUX-DPO and an image generated by FLUX-CPO. Participants are asked to base their comparison on 7 pre-defined root dimensions from a ``Domain-Expert Agent" knowledge as criteria, and to conduct pairwise comparisons on the image pairs in Group G1 and Group G2, respectively. They have to select the superior image from the two under each dimension (7 comparisons in total).

We recruit a total of 10 participants, with an age distribution between 20 and 30. All participants have (or are pursuing) a professional background in art or design, ensuring they possess the professional judgment ability for the aesthetic standards of oil paintings.

The results of the user study are shown in \cref{fig:user_study}.
The data shows that in the SDXL-based comparison (G1 group), 63.5\% of the preference is given to the images generated by our SDXL-CPO+NPO. In the FLUXbased comparison (G2 group), the FLUX-CPO method obtain a user preference as high as 84.1\%.

Whether based on the SDXL or FLUX base model, our CPO achieve a significantly higher user preference rate in direct comparison with the DPO baseline. This result strongly proves that our proposed method has significant superiority in optimizing complex human preferences and enhancing the subjective perceptual quality of generated images.

\textbf{Notice to Human Subjects.}
We issued a notice to subjects to inform them of data collection and use before the experiment:

\begin{quote}
    ``Dear volunteers, thank you for your support of our research. We are researching an image generation algorithm based on Complex Preference Optimization (CPO) and applying it to the generation of oil paintings. All information related to your participation in the study will be displayed in the research records. All information will be processed and stored according to local laws and policies on privacy. Your name will not appear in the final report. When mentioning the data you provide, only the individual number assigned to you will be mentioned. We respect your decision whether to volunteer for this study. If you decide to participate in this study, you can sign this informed consent form.''
\end{quote}

The use of user data has been approved by the Institutional Review Board of the primary author's institution.

\section{More Qualitative Results}
\begin{figure*}[t]
    \centering
    \includegraphics[width=0.9\linewidth]{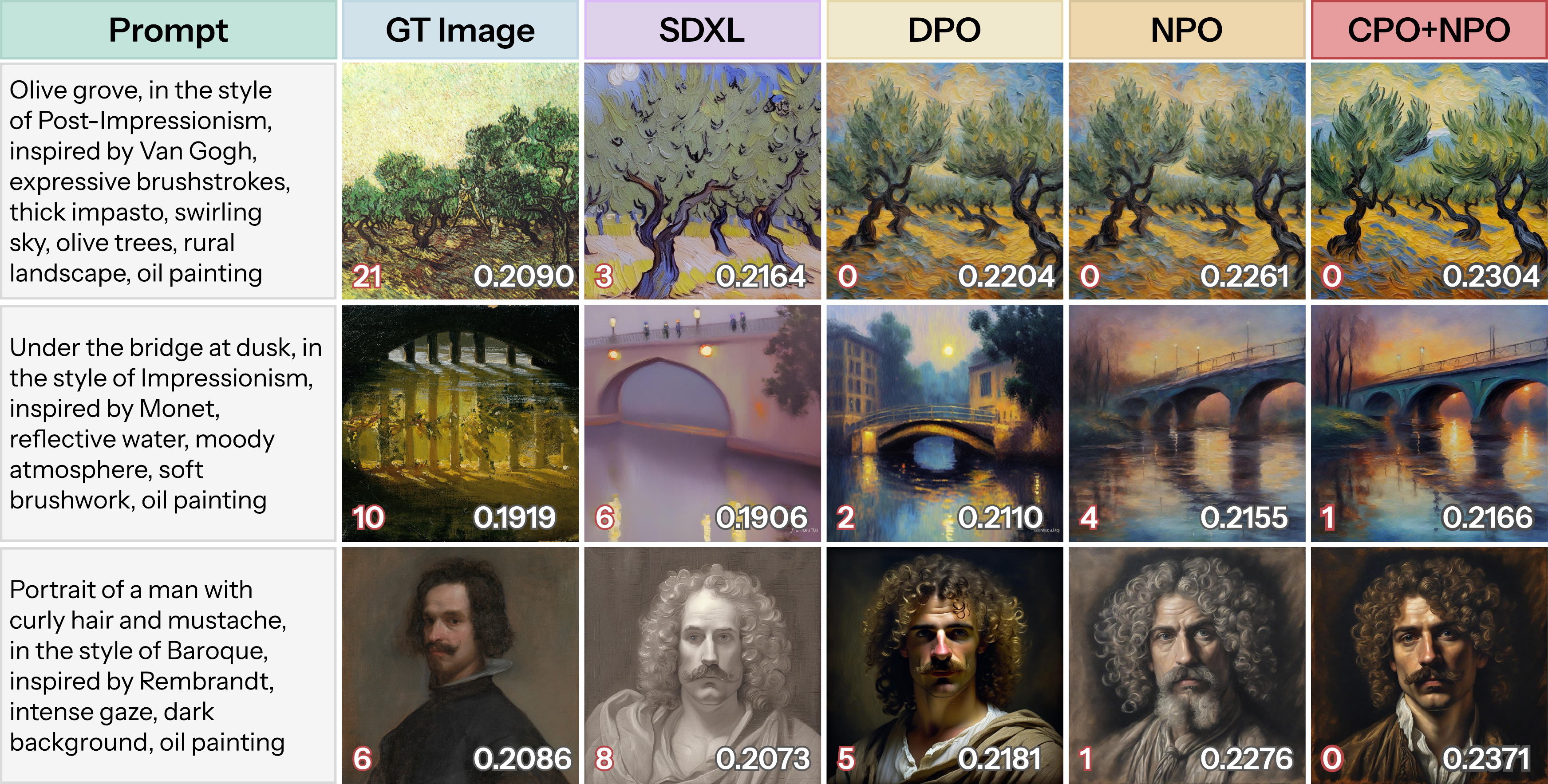} 
    \caption{Visual comparison of different baselines and our CPO. \#A\_neg ($\downarrow$) and PickScore ($\uparrow$) are annotated in the lower-left and lower-right corners of each image, respectively. CPO outperforms all baselines in both negative-attribute avoidance and preference scoring.}
    \label{fig:qualitative_more}
\end{figure*}

\textbf{Qualitative Results of CPO.} 
\cref{fig:qualitative_more} shows the visual performance of different training methods in artistic style generation tasks, including SDXL, DPO, NPO, and CPO combined with NPO (CPO+NPO). The results indicate that CPO+NPO consistently produces the fewest negative attributes across all examples. CPO+NPO also achieves the highest PickScore, clearly outperforming baseline methods. CPO produces images with more natural, precise brushwork, light and shadow, and style consistency, particularly in the swirling sky of Van Gogh's style, the halo effect in Monet's night scene, and the dramatic lighting in the Baroque portrait.

\begin{figure*}[t]
    \centering
    \includegraphics[width=0.8\linewidth]{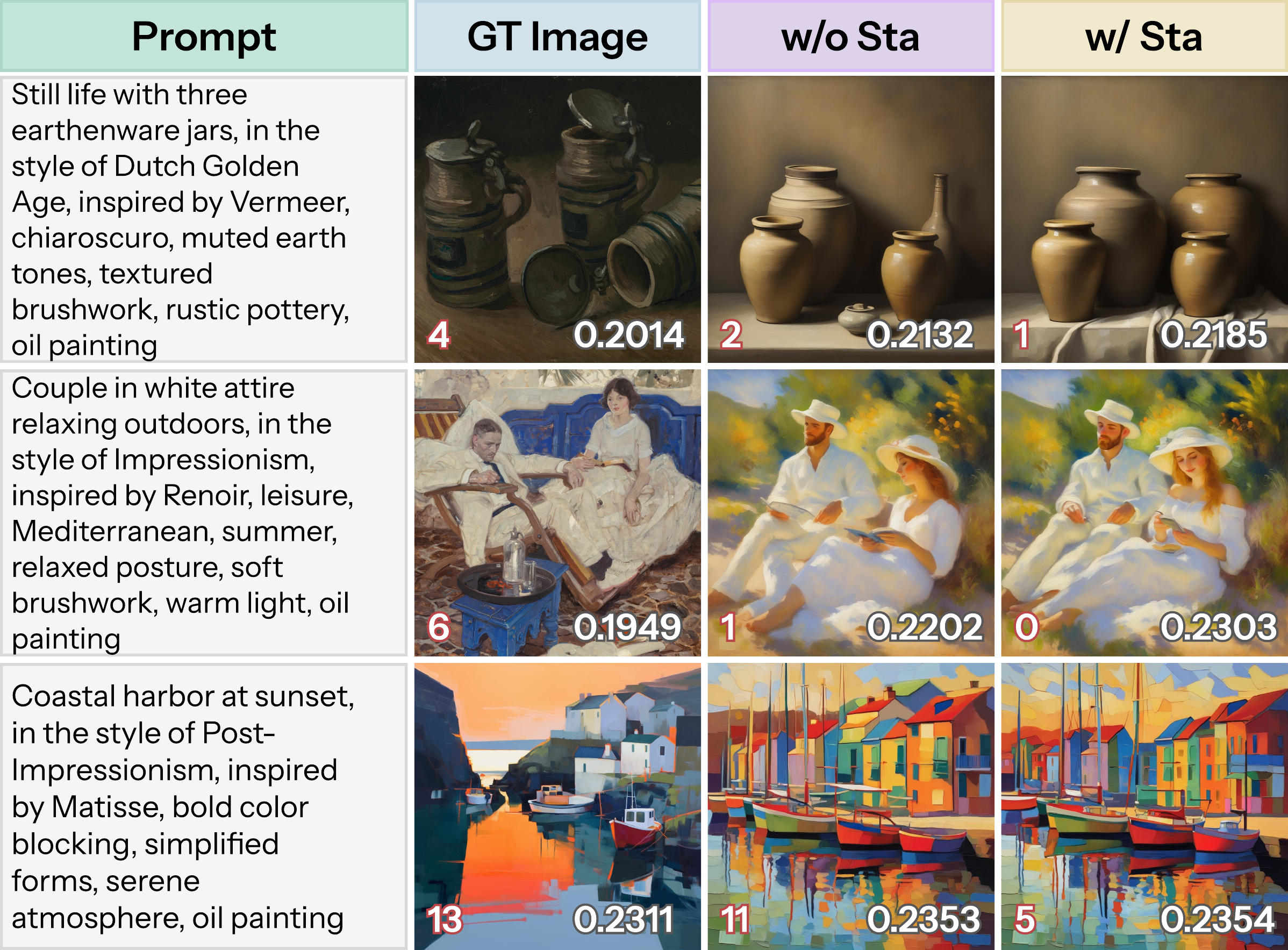} 
    \caption{Visual comparison of different baselines and our CPO. \#A\_neg ($\downarrow$) and PickScore ($\uparrow$) are annotated in the lower-left and lower-right corners of each image, respectively. CPO outperforms all baselines in both negative-attribute avoidance and preference scoring.}
    \label{fig:qualitative_stabilization}
\end{figure*}

\textbf{Qualitative Results of Stabilization Strategy.} \cref{fig:qualitative_stabilization} shows the effect of the stabilization strategy before and after implementation. The results show that the strategy reduces negative attributes across all examples. The stabilization strategy also improves the overall PickScore. In terms of details, the still life shows more coherent light and shadow, the Impressionist figure has better harmony in lighting and skin tone, and the Post-Impressionism harbor displays more stable color blocks and water reflections.

\section{Additional Explanation on Stabilization Strategy}
\label{sec:stabilization_strategy}
\subsection{Effectiveness Analysis}
\begin{figure}[t]
    \centering
    \begin{subfigure}[b]{0.45\textwidth}
        \includegraphics[width=1\linewidth]{sec/figures/loss_curve.pdf} 
        \subcaption{Visualization of the winning and losing parts of the loss function.}
    \end{subfigure}
    \begin{subfigure}[b]{0.45\textwidth}
        \includegraphics[width=1\linewidth]{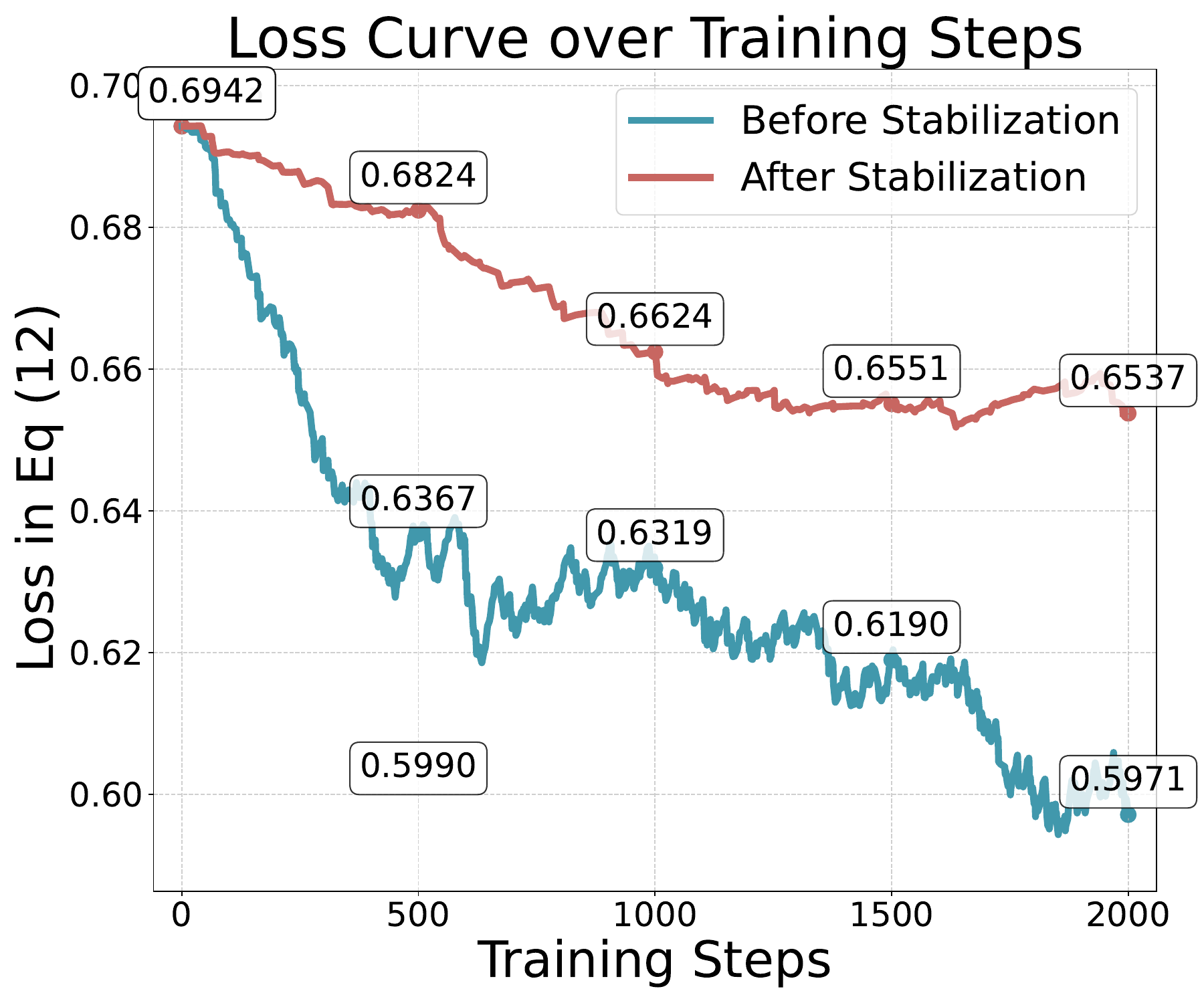}
        \subcaption{Visualization of the overall trend of the loss function.}
    \end{subfigure}
    
    \caption{Curves of the separated winning and losing parts of the loss function, together with the overall loss trend, under the with- and without-stabilization settings over training steps. The loss used in (b) corresponds to Eq(12) in the main paper.}
    \label{fig:sm-loss-curve}
\end{figure}

To assess the effectiveness of our stabilization strategy, we visualize the evolution of the winning term, the losing term, and the overall loss over training steps under both the with- and without-stabilization settings, as shown in \cref{fig:sm-loss-curve}. 

\cref{fig:sm-loss-curve} (b) shows the gap between the positive (winning) and negative (losing) parts. Unlike conventional DPO-style objectives that intentionally enlarge this margin, our method does not aggressively push the positive–negative separation and instead adopts a more balanced and stable approach. 

Classical DPO explicitly aims to maximize this margin, but doing so often comes at the cost of degrading the model’s fit on both positive and negative samples, as illustrated by the blue curves in \cref{fig:sm-loss-curve} (a), thereby sacrificing the model’s learning behavior on desirable positive samples. 
In contrast, we argue that the optimization should also account for how well the model fits the positive samples. As shown by the red curves in \cref{fig:sm-loss-curve} (a), our stabilized training achieves a noticeably lower winning-term loss, indicating stronger learning of positive attributes. 

Ideally, the optimization should move in a direction where the model improves its fit on positive samples while deteriorating its fit on negative samples. Although our method represents a meaningful step toward this objective, it does not yet fully achieve this ideal separation. We regard this as an important direction for future work.


\subsection{Gradient Analysis}
To theoretically justify the effectiveness of our stabilization strategy, we analyze the gradient behavior of the proposed objective. 
Let $\mathcal{L}_{win} = \|\boldsymbol{z}^w-\boldsymbol{\epsilon}_\theta(\boldsymbol{x}_t,t)\|_2^2$ and $\mathcal{L}_{lose} = -\|\boldsymbol{z}^l-\boldsymbol{\epsilon}_\theta(\boldsymbol{x}_t,t)\|_2^2$ denote the winner and loser terms in the original CPO objective, respectively.
The gradient of the original loser term with respect to the model output $\boldsymbol{\epsilon}_\theta$ is derived as:
\begin{equation}
    \nabla_{\boldsymbol{\epsilon}_\theta} \mathcal{L}_{lose} = -2(\boldsymbol{\epsilon}_\theta - \boldsymbol{z}^l),
\end{equation}
which directs the optimization to push $\boldsymbol{\epsilon}_\theta$ away from the negative prototype $\boldsymbol{z}^l$.
However, its magnitude $\|\nabla_{\boldsymbol{\epsilon}_\theta} \mathcal{L}_{lose}\|_2 = 2\|\boldsymbol{\epsilon}_\theta - \boldsymbol{z}^l\|_2$ grows unbounded as the model successfully unlearns the negative attributes, leading to gradient dominance over the winner term.

\begin{figure}[t]
    \centering
    \includegraphics[width=1\linewidth]{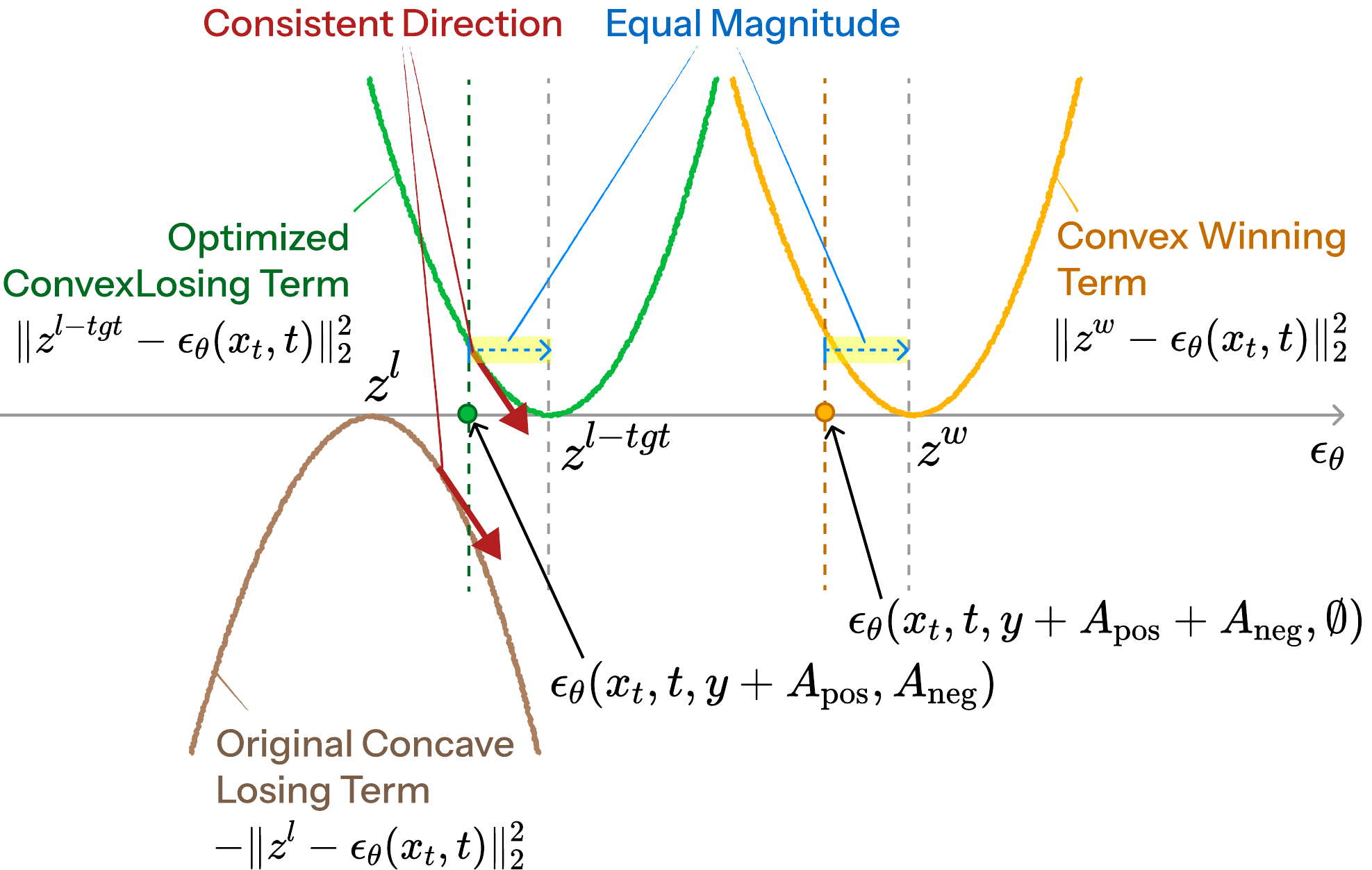} 
    \caption{Illustration of the function transformation in the stabilization strategy. It transforms the originally concave losing term into an equivalent convex formulation. The transformed term preserves the direction of the original losing term, but its optimization magnitude is matched to that of the winning term, ensuring stability during training.}
    \label{fig:sm-function-figure}
\end{figure}

In our stabilized objective $L_{CPO-S}$, we introduce the surrogate target $\boldsymbol{z}^{l-tgt}$. 
Treating $\boldsymbol{z}^{l-tgt}$ as a fixed target (via stop-gradient), the gradient of the new loser term $\mathcal{L}_{stab} = \|\boldsymbol{z}^{l-tgt}-\boldsymbol{\epsilon}_\theta\|_2^2$ is:
\begin{equation}
    \nabla_{\boldsymbol{\epsilon}_\theta} \mathcal{L}_{stab} = -2(\boldsymbol{z}^{l-tgt} - \boldsymbol{\epsilon}_\theta).
\end{equation}
Substituting the definition of $\boldsymbol{z}^{l-tgt} = \boldsymbol{\epsilon}_\theta + \frac{\boldsymbol{\epsilon}_\theta - \boldsymbol{z}^l}{\|\boldsymbol{\epsilon}_\theta - \boldsymbol{z}^l\|_2} \|\boldsymbol{\epsilon}_\theta - \boldsymbol{z}^w\|_2$, we obtain:
\begin{equation}
    \begin{aligned}
    \nabla_{\boldsymbol{\epsilon}_\theta} \mathcal{L}_{stab} &= -2 \left( \frac{\boldsymbol{\epsilon}_\theta - \boldsymbol{z}^l}{\|\boldsymbol{\epsilon}_\theta - \boldsymbol{z}^l\|_2} \|\boldsymbol{\epsilon}_\theta - \boldsymbol{z}^w\|_2 \right) \\
    &= -2 \cdot \underbrace{\frac{\boldsymbol{\epsilon}_\theta - \boldsymbol{z}^l}{\|\boldsymbol{\epsilon}_\theta - \boldsymbol{z}^l\|_2}}_{\text{Direction}} \cdot \underbrace{\|\boldsymbol{\epsilon}_\theta - \boldsymbol{z}^w\|_2}_{\text{Magnitude}}.
    \end{aligned}
\end{equation}
This derivation reveals two critical properties as shown in \cref{fig:sm-function-figure}:
(1) \textbf{Directional Consistency:} The gradient direction aligns with $-(\boldsymbol{\epsilon}_\theta - \boldsymbol{z}^l)$, which is identical to the original repulsive force in $\mathcal{L}_{lose}$, ensuring the model continues to unlearn negative attributes.
(2) \textbf{Magnitude Normalization:} The gradient norm is rescaled to $2\|\boldsymbol{\epsilon}_\theta - \boldsymbol{z}^w\|_2$. This explicitly matches the magnitude of the winner term's gradient $\|\nabla_{\boldsymbol{\epsilon}_\theta} \mathcal{L}_{win}\|_2$, guaranteeing a balanced optimization landscape throughout the training process.

\section{Ablation Study of the Dynamic Process Reward Parameter \texorpdfstring{$\omega$}{omega}}
We conduct an ablation study on the guidance strength hyperparameters $\omega_w$ and $\omega_l$ in CPO. 
For simplicity, we set $\omega_w = \omega_l = \omega$ and evaluate $\omega \in \{1.0, 1.5, 2.0, 2.5, 3.0\}$ 
on the same test set, keeping all other hyperparameters fixed. 
Evaluation metrics are identical to those in the main paper: $\#A_{\text{neg}}$, FID ~\cite{NEURIPS2023_dd83eada_FID}, 
PickScore~\cite{Yuval2023Pick-a-Pic}, HPSv2 ~\cite{wu2023hpsv2}, ImageReward ~\cite{xu2024ImageReward}, and Aesthetic Score~\cite{Schuhmann2022LAION_5B}. 
Results are shown in Table~\ref{tab:ablation-omega}.

Experimental results indicate that as $\omega$ increases from 1.0 to 3.0, the average number of negative attributes in the generated images decreases monotonically from 5.26 to 4.87, confirming that enhanced guidance strength effectively suppresses the generation of negative attributes. However, the FID increases from 86.61 to 90.18, indicating that excessively strong guidance may impair the visual quality of the generated images. In terms of human preference evaluation, ImageReward and Aesthetic scores show continuous improvement with increasing $\omega$, while PickScore and HPSv2 achieve an optimal balance at $\omega=2.0$. Considering the trade-off between negative attribute suppression and visual quality preservation, we ultimately select $\omega=2.0$ as the default parameter, which achieves the best balance between the number of negative attributes (5.18) and multiple human preference metrics.

\begin{table}
  \caption{Ablation study under different \texorpdfstring{$\omega$}{omega}. AN, PS, IR, and LA denote \#A\_neg, PickScore, ImageReward, and LAION-Aesthetic.}
  \label{tab:ablation-omega}
  \centering
  \footnotesize
  \begin{tabular}{@{}lcccccc@{}}
    \toprule
    
    \textbf{$\omega$} & \textbf{\#AN}$^{\downarrow}$ & \textbf{FID}$^{\downarrow}$ & \textbf{PS}$^{\uparrow}$ & \textbf{HPS}$^{\uparrow}$ & \textbf{IR}$^{\uparrow}$ & \textbf{LA}$^{\uparrow}$ \\
    \midrule
    1.0 & 5.260 & \textbf{86.61} & \textbf{0.2179} & 0.2819 & 0.9064 & 6.575 \\
    1.5 & 5.230 & 87.25 & 0.2172 & 0.2837 & 0.9133 & 6.584 \\
    2.0 & 5.180 & 87.37 & 0.2083 & \textbf{0.3039} & 0.9312 & 6.581 \\
    2.5 & 4.940 & 88.91 & 0.2171 & 0.2865 & 0.9367 & 6.599 \\
    3.0 & \textbf{4.870} & 90.18 & 0.2170 & 0.2871 & \textbf{0.9437} & \textbf{6.602} \\
    
    \bottomrule
  \end{tabular}
\end{table}

\section{Additional Details on CPO}
\subsection{Trajectory Description}


Further elaborating on Section 5.2, our Complex Preference Optimization (CPO) objective fundamentally addresses a core computational difficulty faced by standard Direct Preference Optimization (DPO). Methods like DPO attempt to compare the likelihoods $p_\theta(x^w|y)$ versus $p_\theta(x^l|y)$, which involves computing the probabilities over the entire reverse process $p_\theta(x_{1:T}|x_0)$. This calculation is intractable in practice, necessitating approximations by the forward $q_\theta(x_{1:T}|x_0)$ that introduce inherent errors and inefficient training. As illustrated in \cref{fig:trajectory}, CPO circumvents this by operating in the latent space and leveraging the auxiliary model $\theta_1$ to construct deterministic and controllable preference trajectories. 
This does not necessarily imply a smaller propagation error, but the error becomes controllable and exploitable, thereby enabling more efficient training.
For any given real image $x_0 \in \mathcal{D}$ and its prompt $y$, the image is first diffused to a shared noisy state $x_t$. From this identical starting point $x_t$, our method deterministically samples two reverse trajectories: the positive trajectory $z^w_{1:T}$ and the negative trajectory $z^l_{1:T}$. The positive trajectory is guided by the ideal fine-grained condition ($y$ and $A_{pos}$), while the negative trajectory is guided by the undesirable state ($y$, $A_{pos}$, and $A_{neg}$), representing the attributes we aim to suppress. The central advantage of this construction is that both trajectories are precisely engineered to reconstruct at the same noisy state $x_t$. This shared starting point $x_t$ ensures that CPO focuses its optimization effort precisely on the diverging steps immediately following $x_t$, providing a deterministic and explicit positive or negative gradient at every time step $t$.
This contrasts sharply with original DPO, which only utilizes the final endpoints $x_0^w$ and $x_0^l$, leaving the intermediary trajectory random and intractable, thereby relying on approximations that inherently introduce uncertainty and inefficiency.

\subsection{Mathematical Derivations}
Diffusion-DPO~\cite{wallace2024diffusionDPO} adapts the Direct Preference Optimization (DPO)~\cite{rafailov2023directDPO} framework to the text-to-image diffusion models. The core challenge lies in the intractability of the conditional distribution $p_{\theta}(x_0|c)$ in diffusion models, where $x_0$ is the final generated image and $c$ is the text prompt. This is because $p_{\theta}(x_0|c)$ requires marginalizing over all possible diffusion paths $x_{1:T}$. To address this, Diffusion-DPO leverages the Evidence Lower Bound (ELBO) and reformulate the problem to operate on the full diffusion path $x_{0:T} = (x_0, x_1, \dots, x_T)$. This leads to a new training objective:
\begin{equation}
\label{eq:diffusion-dpo}
\begin{aligned}
L_\text{Diffusion-DPO}= -\mathbb{E}_{(x_0^w, x_0^l) \sim \mathcal{D}} \log \sigma \bigg( \
\beta \mathbb{E}_{\substack{x_{1:T}^w \sim p_\theta(x_{1:T}^w|x_0^w) \\ x_{1:T}^l \sim p_\theta(x_{1:T}^l|x_0^l)}}
\Big[ \\log \frac{p_\theta(x_{0:T}^w)}{p_\text{ref}(x_{0:T}^w)} - \log \frac{p_\theta(x_{0:T}^l)}{p_\text{ref}(x_{0:T}^l)} \Big] \bigg).
\end{aligned}
\end{equation}

The loss in \cref{eq:diffusion-dpo} remains intractable due to the expectation over the reverse process $p_{\theta}(x_{1:T}|x_0, c)$, which involves untrainable path variables. To achieve efficient gradient-based optimization, we make a key approximations. Specifically, we substitute the intractable reverse process $p_{\theta}(x_{1:T})$ with the tractable deterministic trajectories $p_{\theta_1}(x_{1:T})$. As shown in \cref{fig:trajectory} (a), given the noise $z_t$ at the current timestep $t$, we can derive the predicted $\hat{x}_0$ according to the principles of diffusion models:
\begin{equation}
    \hat{x}_0 = \frac{1}{\sqrt{\alpha_t}} \left( x_t - \sigma_t \cdot z_t \right).
\end{equation}
Given $\hat{x}_0$ and $z_t$, we can reconstruct $x_t$ exactly, thereby making the trajectory $p_{\theta_1}(x_{1:T})$ accessible.
\begin{equation}
    x_t = \sqrt{\alpha_t} \hat{x}_0 + \sigma_t z_t
\end{equation}
By applying this approximation and substituting the log-likelihood ratio with the KL-divergence between the $p_{\theta}(x_{1:T})$ and $p_{\theta_1}(x_{1:T})$, the loss simplifies to:
\begin{equation}
\begin{aligned}L(\theta)=- & \mathbb{E}_{ t \sim \mathcal{U}(0, T), x_{t}^{w} \sim p_{\theta_1}\left(x_{t}^{w} \mid \hat{x_{0}^{w}}\right), x_{t}^{l} \sim p_{\theta_1}\left(x_{t}^{l} \mid \hat{x_{0}^{l}}\right)} \\& \log \sigma(-\beta T( \\& +\mathbb{D}_{\mathrm{KL}}\left(p_{\theta_1}\left(x_{t-1}^{w} \mid x_{t}^{w},\hat{x_{0}^{w}}\right) \| p_{\theta}\left(x_{t-1}^{w} \mid x_{t}^{w}\right)\right) \\& -\mathbb{D}_{\mathrm{KL}}\left(p_{\theta_1}\left(x_{t-1}^{w} \mid x_{t}^{w},\hat{x_{0}^{w}}\right) \| p_{\mathrm{ref}}\left(x_{t-1}^{w} \mid x_{t}^{w}\right)\right) \\& -\mathbb{D}_{\mathrm{KL}}\left(p_{\theta_1}\left(x_{t-1}^{l} \mid x_{t}^{l},\hat{x_{0}^{l}}\right) \| p_{\theta}\left(x_{t-1}^{l} \mid x_{t}^{l}\right)\right) \\& \left.+\mathbb{D}_{\mathrm{KL}}\left(p_{\theta_1}\left(x_{t-1}^{l} \mid x_{t}^{l},\hat{x_{0}^{l}}\right) \| p_{\mathrm{ref}}\left(x_{t-1}^{l} \mid x_{t}^{l}\right)\right)\right) .\end{aligned}
\end{equation}
Here we adopt the same strategy as diffusion-DPO~\cite{wallace2024diffusionDPO}, using a uniformly sampled step $t \sim \mathcal{U}(0, T)$. Finally, substituting the definitions of the KL-divergence for diffusion models, which relates to the mean-squared error (MSE) of the predicted noise $\epsilon_{\theta}$7, the final objective for CPO is derived:
\begin{equation}
\label{eq: loss-cpo-SM}
    \begin{aligned}
    L_{CPO}(\theta) =-\mathbb{E}_{t\sim\mathcal{U}(0,T),z_t^w,z_t^l}
     \log\sigma(-\beta T\omega(\lambda_t)( \\
      \|z^w-\epsilon_\theta(x_t,t)\|_2^2-\|z^w-\epsilon_\mathrm{ref}(x_t,t)\|_2^2 \\
      -(\|z^l-\epsilon_\theta(x_t,t)\|_2^2-\|z^l-\epsilon_\mathrm{ref}(x_t,t)\|_2^2)))
    \end{aligned}
\end{equation}
where $z_t^w$ and $z_t^l$ are the noise sampled from the pre-trained expert model $\theta_1$, $\lambda_t$ is the signal-to-noise ratio, and $\omega(\lambda_t)$ is a weighting function (often constant).

This final loss function (\cref{eq: loss-cpo-SM}) directly optimizes the denoising model $\epsilon_{\theta}$ to reduce the noise prediction error for the positive noise ($z_t^w$) relative to the reference model $\epsilon_{ref}$, and conversely, to increase the error for the negative noise ($z_t^l$). The term $\beta T \omega(\lambda_t)$ acts as a dynamic coefficient scaling the preference score.



\section{Hallucination in Agent Behaviors}
To investigate the accuracy of the automatically annotated dataset, we conduct a human verification. We randomly select 100 samples with complex annotations from the original dataset. A total of 10 participants are invited, with a gender ratio of 1:1 and ages ranging from 20 to 30. 

Participants are required to examine all positive and negative attributes across 7 dimensions for each image and record the attributes that actually appeared to calculate the annotation accuracy and verify the reliability of the automatic annotation results. The calculation is defined as follows:
\begin{equation}
    \text{Accuracy} = \frac{\text{Actual Occurrences}}{\text{Occurrences in Annotations}}\times 100\%
    \label{eq:accuracy_formula}
\end{equation}

As shown in \cref{tab:human_verification}, the overall accuracy is 88.71\%. The accuracies for individual dimensions are as follows: Color Relationship (96.18\%), Perspective and Space (91.93\%), Edge Relationship (91.44\%), Light and Shadow (94.49\%), Brushwork and Texture (89.29\%), Composition (88.93\%), and Shape and Form (81.96\%). 

Although a high level of accuracy has been achieved, there remains a slight deviation compared to human judgment. On the one hand, human interpretations of aesthetic attributes inherently involve a certain subjectivity, making complete consensus difficult and potentially affecting labeling accuracy. On the other hand, we believe this deviation does not hinder our task construction or algorithmic optimization. Since our proposed CPO method is designed to encourage the model to generate samples exhibiting positive attributes while suppressing those with negative attributes, accurately identifying positive and negative attributes is more critical than achieving exhaustive annotation coverage.

\begin{table}[!ht]
    \centering
    \caption{Verification of annotation accuracy across 7 dimensions. The results are compared against human judgment, with an overall accuracy of 88.71\%.}
    \label{tab:human_verification}
    \begin{tabular}{lc}
        \toprule
        \textbf{Dimension} & \textbf{Accuracy (\%)} \\
        \midrule
        Color Relationship & 96.18 \\
        Perspective and Space & 91.93 \\
        Edge Relationship & 91.44 \\
        Light and Shadow & 94.49 \\
        Brushwork and Texture & 89.29 \\
        Composition & 88.93 \\
        Shape and Form & 81.96 \\
        \midrule
        \textbf{Overall} & \textbf{88.71} \\
        \bottomrule
    \end{tabular}
\end{table}
\section{Reliability of the SFT Model}
To evaluate the reliability of the model after first-stage SFT training, we test its performance metrics and IoU scores for $A_{pos}$ and $A_{neg}$ predictions under three different inference strategies. Specifically, we compared: Configuration A (the method for first-stage CPO alignment, placing description $y$ and $A_{pos}$ in the prompt and $A_{neg}$ in the negative prompt), Configuration B (placing only $y$ and $A_{pos}$ in the prompt), and Configuration C (placing $y$, $A_{pos}$, and $A_{neg}$ all in the prompt). Detailed results are presented in ~\cref{tab:ABC}. All three configurations achieve high IoU for $A_{pos}$ and low IoU for $A_{neg}$, indicating that after SFT, the model can effectively encode $A_{pos}$ while suppressing the expression of $A_{neg}$, ultimately generating images that accurately reflect the attribute requirements in the prompt,  demonstrating the reliability of SFT. Notably, Configuration A yields the highest $A_{pos}$ IoU, the lowest $A_{neg}$ IoU, and the best overall performance, corroborating the superiority of our CPO approach.

\begin{table}
  \caption{Quantitative evaluation of our SFT-trained model under three prompting configurations. $\text{IoU}_{\text{pos}}$, $\text{IoU}_{\text{neg}}$, PS, HPS, IR, and LA denote IoU scores for $A_{\text{pos}}$ and $A_{\text{neg}}$, PickScore, HPSv2, ImageReward, and LAION-Aesthetic Score.}
  \label{tab:ABC}
  \centering
  \footnotesize
  \setlength{\tabcolsep}{2.2pt}
  \begin{tabular}{@{}lccccccc@{}}
    \toprule
    \textbf{Config} & \textbf{$\text{IoU}_{\text{pos}}$}$^{\uparrow}$ & \textbf{$\text{IoU}_{\text{neg}}$}$^{\downarrow}$ & \textbf{FID}$^{\downarrow}$ & \textbf{PS}$^{\uparrow}$ & \textbf{HPS}$^{\uparrow}$ & \textbf{IR}$^{\uparrow}$ & \textbf{LA}$^{\uparrow}$ \\
    \midrule
    \textbf{A} & \textbf{0.7780} & \textbf{0.3928} & \textbf{88.3168} & \textbf{0.1939} & \textbf{0.2592} & \textbf{0.4843} & \textbf{6.1051} \\
    B & 0.6617 & 0.3975 & 91.3259 & 0.1912 & 0.2467 & 0.4342 & 6.0487 \\
    C & 0.6539 & 0.4253 & 93.0775 & 0.1925 & 0.2551 & 0.4462 & 6.0599 \\
    \bottomrule
  \end{tabular}
\end{table}

\section{Negative Noise Construction}
Here, we clarify why the direction of our negative noise guidance is derived from $(y, A_{pos}, A_{neg})$ rather than solely from $A_{neg}$. In our domain-specific fine-grained evaluation , each image is first annotated with its corresponding positive attributes based on the content. However, when an image exhibits local deficiencies, certain positive attributes may not be properly realized; in such cases, the image is additionally annotated with the corresponding negative attributes. In other words, the positive labels encode the complete attribute information of an image, whereas the negative labels only identify which aspects are deficient. 

For example, if $A_{pos}$ includes a compositional attribute such as “circular composition,” then the associated negative attribute would be “absence of shape-breaking elements,” since circular composition intrinsically requires such elements. If we were to provide only the negative label “aabsence of shape-breaking elements” without the accompanying compositional information, the semantics would be incomplete.



\section{Differences from and Advantages over Inversion-Based DPO}
Our proposed Complex Preference Optimization (CPO) framework significantly advances diffusion model alignment beyond existing inversion-based DPO methods, such as DDIM-InPO (InPO)~\cite{lu2025inpo} and Inversion-DPO~\cite{li2025inversionDPO}, offering key advantages rooted in signal granularity, training efficiency, and optimization stability. The primary distinction lies in the granularity of the alignment signal: existing inversion-based DPO approaches fundamentally rely on maximizing monolithic, coarse preference (binary winner/loser pairs). In contrast, CPO introduces a novel, domain-specific evaluation criterion that is hierarchical, multi-dimensional, discrete, and non-equilibrium, allowing it to explicitly decouple positive ($A_{pos}$) and negative ($A_{neg}$) attributes within a single sample. This attribute decoupling enables fine-grained guidance, steering the model toward desired characteristics while actively suppressing undesirable ones, a capability absent in methods optimizing only for a simple preference score or implicit reward derived from inversion.

Furthermore, CPO exhibits superior computational efficiency and enhanced training stability. While Inversion-DPO leverages DDIM inversion to achieve a more precise approximation of the diffusion path compared to Diffusion-DPO and InPO is highly efficient, aiming for state-of-the-art performance in just 400 training steps, CPO offers compelling practical speed gains. For instance, achieving stable convergence for one epoch on the SDXL model with CPO requires approximately 10 GPU hours, representing a significant reduction in overhead even compared to optimized inversion-based methods, which, in practice, may require around 138 GPU hours for a comparable epoch (Inversion-DPO reports acceleration factors greater than $2\times$ over Diffusion-DPO). Additionally, CPO addresses a critical instability inherent in the DPO objective itself by incorporating a novel stabilization strategy $L_{CPO-S}$. This strategy specifically counteracts the imbalance where the concave loss term for losing samples dominates the convex loss term for winning samples, resulting in demonstrably smoother and more robust training convergence than the non-stabilized variant ($L_{CPO}$). In contrast, inversion-based methods focus their stability gains primarily on improving the accuracy of the underlying diffusion process trajectory rather than rectifying this specific gradient entanglement issue in the DPO loss function.

\section{Discussion}
\subsection{The Special Variant of CPO}
CPO is inherently designed to handle multi-dimensional and decoupled preference signals. It is crucial to examine the relationship between CPO and existing methods when its complexity is reduced. If the attribute system within CPO is constrained to a single dimension with one-level deep, the CPO objective effectively simplifies to a form highly similar to the Direct Preference Optimization (DPO)~\cite{wallace2024diffusionDPO}. This is because the core of CPO is built upon optimizing the log-probability difference between the winning and losing samples, an operational structure that mirrors DPO but is adapted for diffusion models via dynamic noise targets ($z^w, z^l$). This observation positions CPO as a generalized preference optimization framework that extends DPO's binary preference capability to complex, multi-criteria alignment signals within generative models. Furthermore, it is important to distinguish CPO from the Binary Classifier Optimization (BCO)~\cite{jung2025binaryBCO} approach. BCO transforms the preference alignment task into a binary classification problem, where a model is trained to classify preferences based on log-probabilities, and the policy is then optimized using the resulting classification logits. In contrast, CPO remains a direct policy optimization method. We do not train an explicit classifier or reward model. Instead, the preference signal is encoded directly into the noise targets, enabling the policy to be updated directly and stably without an auxiliary classification step. This direct preference gradient application differentiates our approach from BCO's classification-mediated optimization strategy.

\subsection{The reliability of CPO}
A key design aspect of our two-stage approach is the reliance on the fine-tuned model $\theta_1$ to generate the dynamic noise targets, $z^w$ (winner) and $z^l$ (loser), used in the CPO objective. A potential critique is that the final model $\theta$ is learning from a surrogate representation of preference---the knowledge learned by $\theta_1$ via Supervised Fine-Tuning (SFT) with attribute prompts---rather than directly from the ground-truth fine-grained attributes $A_{pos}$ and $A_{neg}$ of the original dataset $\mathcal{D}$. We acknowledge this as a limitation stemming from the inherent difficulty of performing direct, stable preference optimization on complex, multi-dimensional, and non-equilibrium signals. However, the utilization of a surrogate model is a common and often necessary practical trick in modern generative modeling and reinforcement learning. For instance, in Generative Adversarial Networks (GANs)~\cite{goodfellow2014generative}, the generator optimizes through gradients provided by the discriminator rather than direct data likelihood. Similarly, diffusion distillation techniques like DisBack~\cite{zhang2025disback} and preference optimization methods like DDO~\cite{zheng2025adversarialDPO} utilize an auxiliary model or a discriminator as a surrogate for knowledge transfer or preference signal. Furthermore, in standard Reinforcement Learning from Human Feedback (RLHF), an explicit reward model is trained from human preference data and subsequently acts as a surrogate during the policy optimization stage. In our work, $\theta_1$ serves as a knowledge-guided surrogate model, injecting and structuring the complex domain expertise such that the decoupled positive and negative attributes can be dynamically translated into quantifiable noise targets $z^w$ and $z^l$. Future research will explore more sophisticated techniques to bypass $\theta_1$ and achieve direct, stable alignment with raw $A_{pos}$ and $A_{neg}$ labels.

\subsection{The Generalizability of CPO}
Another critical point is the generalizability of our domain-specific fine-grained evaluation criteria. We instantiate our approach in the painting generation domain with a 5-level hierarchy, 7 root dimensions, and 246 pairs of attributes. We emphasize that while the content of the attributes is domain-specific (e.g., "Color Relations" and "Brushstroke" for paintings ), the paradigm characterized by being multi-dimensional, discrete, and non-equilibrium is proposed as a universal structure for modeling complex human expertise. The core innovation is in the CPO objective and its ability to process such a rich signal, irrespective of the domain. Our method is designed to be easily extensible to other complex generation scenarios, provided a similar complex criteria.



\section{Failure Cases and Limitation}
\label{sec:failure_cases}
\begin{figure}[t]
    \centering
    \includegraphics[width=1.0\linewidth]{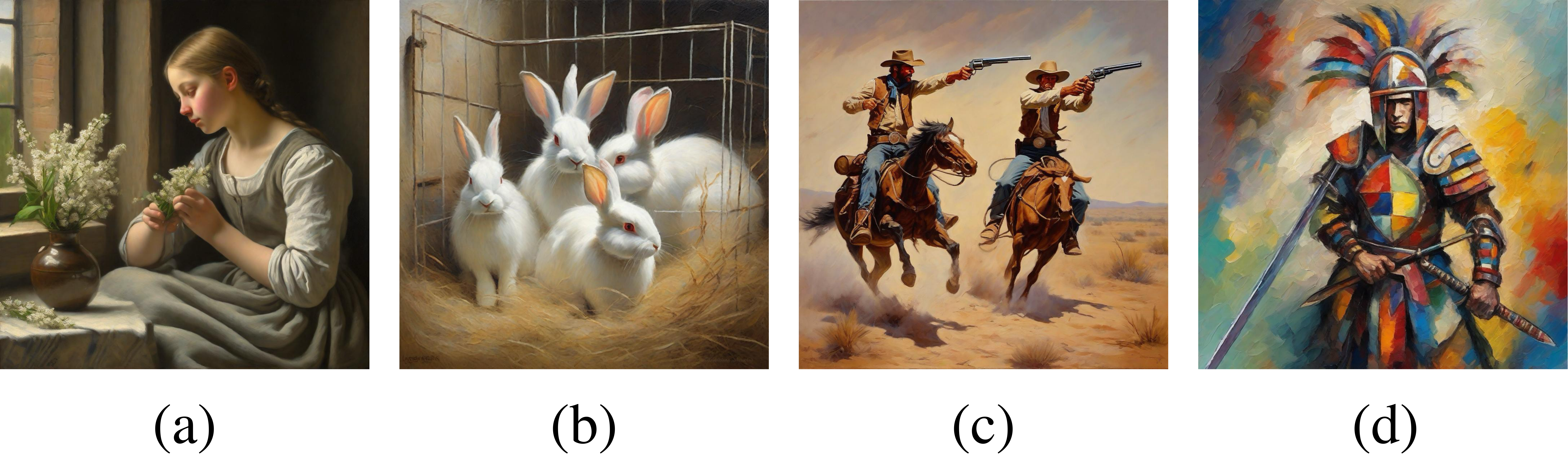} 
    \caption{Additional results of failure examples.}
    \label{fig:failure}
\end{figure}

\textbf{Failure cases.} While CPO can generate high-quality images, it remains constrained by the inherent limitations of the base model, and typical failure modes persist. As shown in the ~\cref{fig:failure}, these mainly include: (a) anatomical structural defects (e.g., finger distortion), (b) quantity errors (e.g., abnormal number of rabbit ears), (c) scale anomalies (e.g., excessively long revolver barrel), and (d) spatial misalignment (e.g., incorrect sword placement). Additionally, some samples fail to satisfy specific positive attribute requirements; for example, (b) does not actually meet the abstract characteristics required by ``abstract geometry''.

\textbf{Limitation.} As discussed in \cref{sec:failure_cases}, CPO's performance remains constrained by the inherent limitations of the base model, occasionally failing to fully satisfy all specified positive attribute requirements. Furthermore, as elaborated in \cref{sec:stabilization_strategy}, while our stabilization strategy enhances positive sample fitting, it has yet to achieve the ideal optimization objective of simultaneously improving positive sample fitting and degrading negative sample fitting. These limitations will be prioritized for exploration and resolution in future work.

\section{Social Impact}
CPO and the underlying hierarchical, fine-grained evaluation criteria present a substantial positive impact on generative AI by enabling models to align with nuanced human expertise, potentially elevating the quality and controllability of generated content in domains like digital art and design. By shifting the alignment paradigm from coarse, binary preference to multi-dimensional, attribute-decoupled criteria, our method facilitates the integration of complex, domain-specific knowledge into generative models, leading to outputs that are more aesthetically sophisticated and technically sound according to expert standards. This advancement can empower creators by providing tools that adhere to higher, more specific quality benchmarks, thereby raising the overall standard of machine-generated content.

However, the technology's effectiveness in instilling expert-defined criteria necessitates consideration of potential risks. The explicit design to favor specific positive attributes $A_{pos}$ and suppress negative ones $A_{neg}$ could inadvertently introduce or amplify biases present in the expert-annotated dataset. If the domain-specific criteria reflect a narrow, culturally or demographically homogenous view of ``good" or ``bad" attributes, the resulting aligned model may exhibit a reduced diversity, potentially marginalizing minority or unconventional styles. Future work must focus on actively diversifying the expert-defined criteria and the corresponding training data to ensure that CPO promotes universally beneficial and equitable generative models, preventing the entrenchment of a single, privileged aesthetic or technical standard.


\end{document}